\documentclass[11pt, a4paper]{googledeepmind}

\pdftrailerid{redacted}

\makeatletter
\renewcommand\bibentry[1]{\nocite{#1}{\frenchspacing\@nameuse{BR@r@#1\@extra@b@citeb}}}
\makeatother
\usepackage{kantlipsum, lipsum}
\usepackage{dsfont}
\usepackage{gdm-colors}
\usepackage{cleveref}
\usepackage[absolute,overlay]{textpos}
\usepackage{booktabs}
\usepackage{array}
\usepackage{multirow}
\usepackage{colortbl}
\usepackage{makecell}
\usepackage[utf8]{inputenc}
\usepackage[T1]{fontenc}
\usepackage{longtable}

\usepackage{fancyhdr}
\usepackage{bm}

\usepackage[numbers, sort&compress]{natbib}

\graphicspath{{figures/}}

\definecolor{LightCyan}{rgb}{0.88,1,1}

\title{MedVersa: A Generalist Foundation Model for Medical Image Interpretation}

\correspondingauthor{Pranav Rajpurkar, PhD {\href{mailto:pranav_rajpurkar@hms.harvard.edu}{(pranav\_rajpurkar@hms.harvard.edu)}}}

\author[$\bigstar$]{Hong-Yu Zhou PhD}
\author[*,$\bigstar$]{Julián Nicolás Acosta MD}
\author[*,$\dag$]{Subathra Adithan MD}
\author[$\ddag$]{Suvrankar Datta MD}
\author[$\triangle$]{Eric J. Topol MD}
\author[$\bigstar$]{Pranav Rajpurkar PhD}

\affil[*]{Equal contributions}
\affil[$\bigstar$]{Department of Biomedical Informatics, Harvard Medical School, Boston, USA}
\affil[$\dag$]{Jawaharlal Institute of Postgraduate Medical Education and Research, Puducherry, IN}
\affil[$\ddag$]{Department of Radiodiagnosis and Interventional Radiology, All India Institute of Medical Sciences (AIIMS), New Delhi, IN}
\affil[$\triangle$]{Scripps Research Translational Institute, Scripps Research, La Jolla, CA, USA.}

\banner{F1}{\textbf{MedVersa} is capable of performing both vision-language and vision-centric medical imaging tasks. It can accept a wide range of inputs, including images of various types (e.g., multimodal, multiview, or longitudinal) and language requests (clinical context and dynamic tasking protocol). The model's optimizable orchestrator, powered by a large language model, evaluates whether to perform the task independently or to incorporate visual modeling components. Combining language models and vision modules, MedVersa leverages both visual and linguistic supervision for training, as well as generates multimodal outputs, showcasing its versatility and potential for real-world applications in medical image interpretation.}

\begin{abstract}
\raggedright
Current medical AI systems are often limited to narrow applications, hindering widespread adoption. We present MedVersa, a generalist foundation model trained on tens of millions of compiled medical instances. MedVersa unlocks generalist learning from multimodal inputs and outputs, representing the first example of a generalist model reaching competitive performance with leading specialized solutions across a variety of medical imaging scenarios. MedVersa achieves state-of-the-art performance in nine tasks, sometimes outperforming counterparts by over 10\%. Radiologist evaluation shows MedVersa-generated reports get superior performance in 95\% of normal studies, while matching or exceeding human reports in 71\% of cases overall. User studies showed notable reductions in report writing time and discrepancies with the use of MedVersa. Our findings underscore the value of flexible, multimodal AI systems in advancing medical image interpretation and supporting clinical expertise.
\end{abstract}

\begin{document}
\maketitle
\newpage
\clearpage

\section{Introduction}
The field of medical artificial intelligence (AI) has been advancing at a rapid pace, ushering in a new era of diagnostic accuracy and patient care. Within this dynamic landscape, researchers have been focusing their efforts on developing solutions for specific tasks, such as identifying chest pathologies \citep{Rajpurkar2017-mh,Wang2017-sg,Irvin2019-hd,Johnson2019-er,Tiu2022-zm} and classifying skin diseases \citep{Liu2020-fv,Esteva2017-bq,Daneshjou2022-tk}. Similarly, the majority of medical AI products approved by the US Food and Drug Administration for clinical use have been designed to address one or two specific tasks \citep{Joshi2022-ex}. However, this task-specific approach may limit the real-world clinical applications of these AI systems, as they may not be able to adapt to the diverse and complex needs of healthcare settings \citep{Moor2023-cf,Rajpurkar2023-bp}.

Addressing this concern, generalist medical artificial intelligence (GMAI) was proposed to utilize recent advances in foundation models \citep{Bommasani2021-qs} for more flexible problem solving \citep{Moor2023-iy}. However, contemporary GMAI models have been designed to learn from natural language supervision \citep{Tu2023-ef,Wu2023-aa,Moor2023-iy,Lu2023-px,Huang2023-kf}. Although these models work well in vision-language tasks, it does not readily apply to a majority of vision-centric problems, such as detection and segmentation, which are indispensable to medical image interpretation \citep{Chen2022-nk,Wang2023-eu,Zhang2022-xc}.

\begin{figure}[t]
    \centering
    \includegraphics[width=\textwidth]{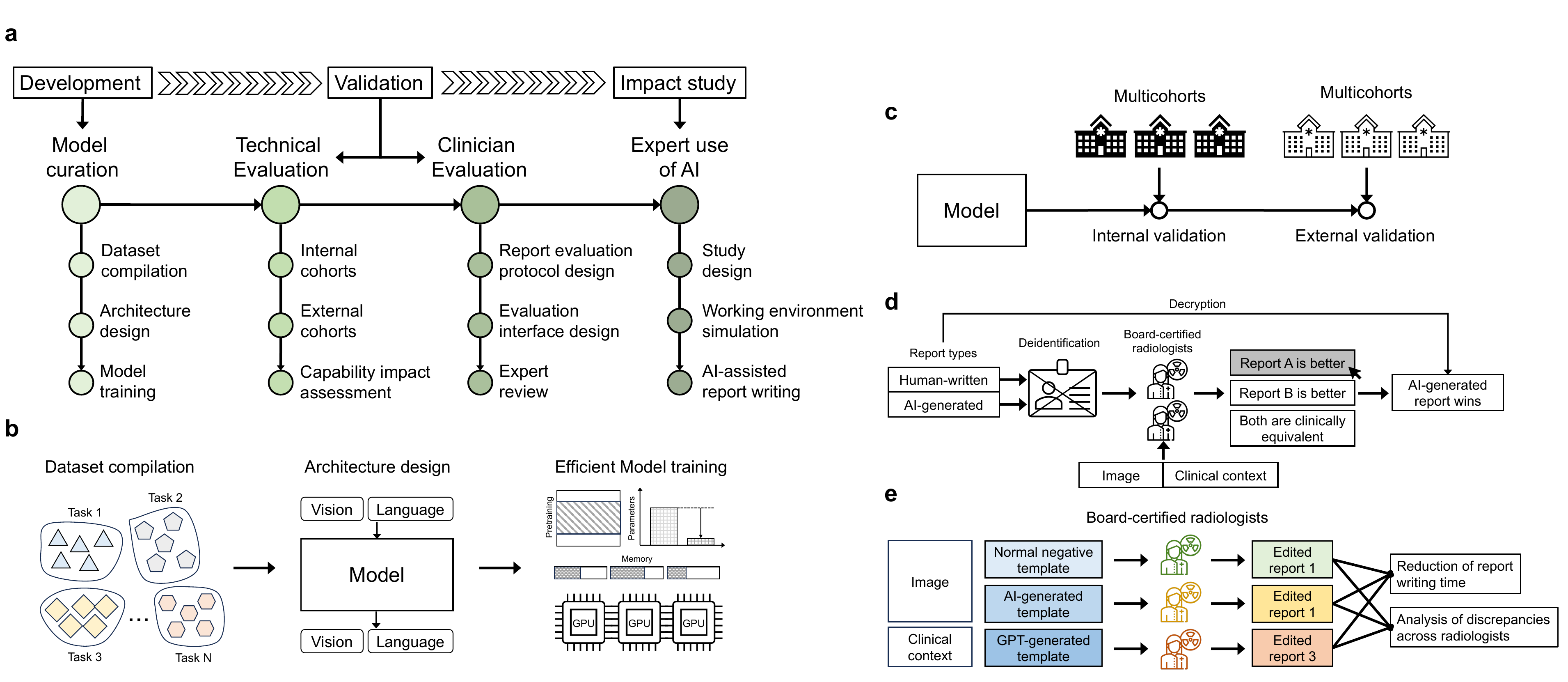}
    \caption{\textbf{Study overview}. \textbf{a,} The three-stage pipeline showing capability development, model validation, and impact study phases. Each phase builds upon model curation, technical and clinical evaluation, and expert use of AI, respectively. \textbf{b,} Development process that outlines dataset compilation across tasks, architecture design integrating multimodal inputs and outputs, and efficient model training strategies. \textbf{c,} Validation framework demonstrating internal and external validation across multicohorts, with emphasis on task improvements through generalist learning and clinical relevance assessment. \textbf{d,} Protocols for comparing human-written and AI-generated reports through blinded assessment by board-certified radiologists to ensure unbiased evaluation. \textbf{e,} Impact study workflow showing the comparative analysis of normal negative template, AI-generated template, and GPT-generated template to assess reduction in report writing time and discrepancies across radiologists.}
    \label{fig:2}
\end{figure}

\begin{figure}[!htp]
    \centering
    \includegraphics[width=\textwidth]{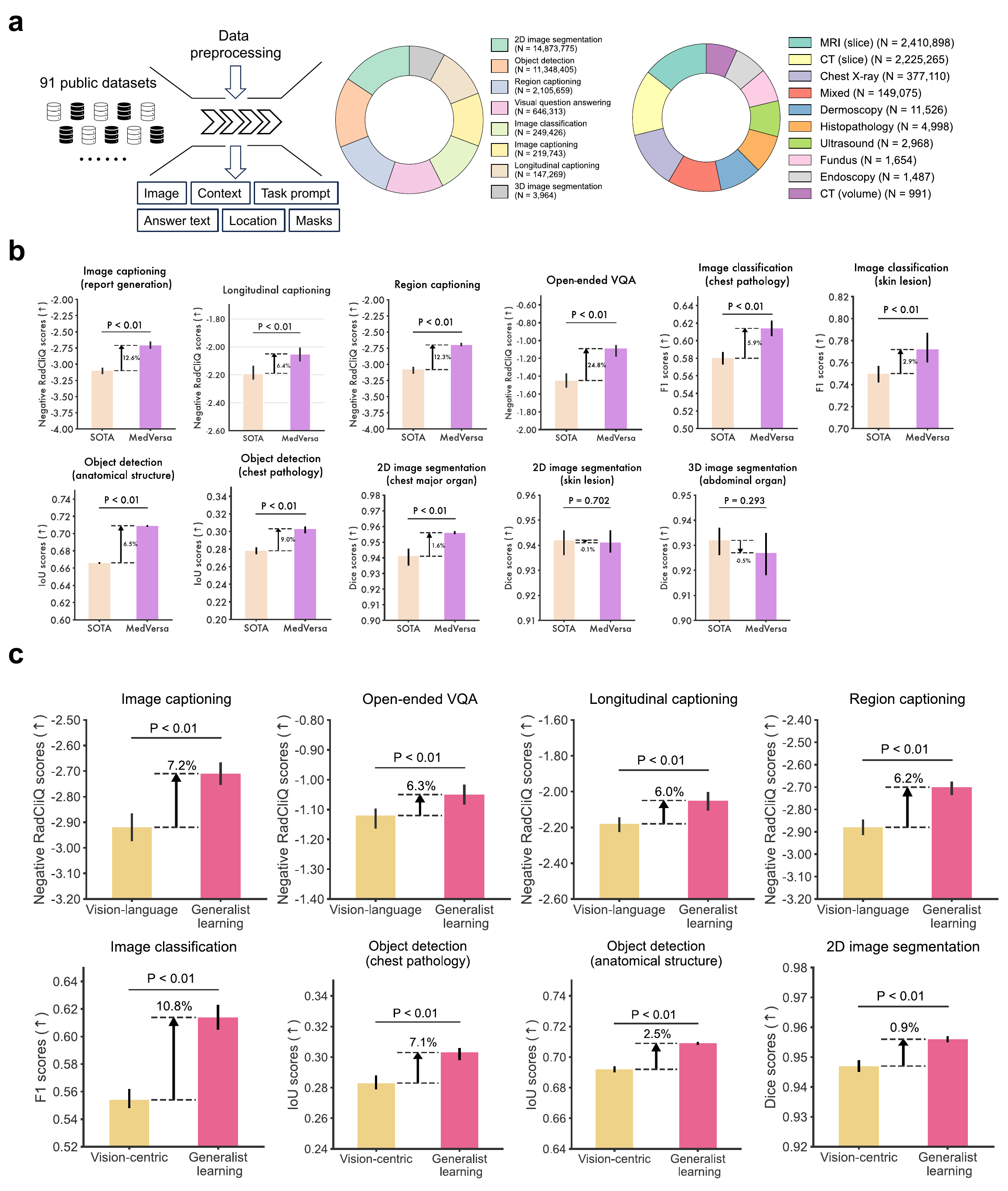}
    \caption{\textbf{Dataset and model performance}. \textbf{a,} Data preprocessing pipeline showing the integration of public datasets and their transformation into varied components including image, context, task prompt, answer text, location, and masks, with a circular chart (values are log-transformed) showing the distribution of different tasks in the dataset. \textbf{b,} Comparisons to state-of-the-art specialized solutions. Relative improvments and p-values are also displayed. \textbf{c,} Performance gains brought by doing generalist learning over using vision-language or vision-centric data for model training. Chest radiographs were chosen for their diverse analytical tasks.}
    \label{fig:3}
\end{figure}

By functioning a large language model as an optimizable orchestrator, MedVersa unlocks generalist learning from multimodal inputs and outputs (Fig. \ref{fig:banner}), distinguishing it from traditional approaches \citep{Tu2023-ef,Wu2023-aa,Moor2023-iy,Lu2023-px,Huang2023-kf}. To develop and validate MedVersa, we proposed a three-stage pipeline for capability development, model validation, and impact study, integrating dataset compilation, multimodal architecture design, efficient training, multicohort validation, expert-blinded report comparisons, and workflow analysis to enhance clinical AI applications and report efficiency (Fig. \ref{fig:2}). To develop MedVersa, we compiled 29 million instances for multifaceted medical image interpretation (Fig. \ref{fig:3}{\color{Magenta}a}). MedVersa is the first foundation model that reaches highly competitive performance with leading task-specific solutions across various medical tasks, often outperforming counterparts by large, significant margins (Fig. \ref{fig:3}{\color{Magenta}b}). It is also the first demonstration that generalist learning across vision and language tasks yields mutual benefits (Fig. \ref{fig:3}{\color{Magenta}c}). In practice, MedVersa surpasses MAIRA \citep{Hyland2023-al} and Med-PaLM M \citep{Tu2023-ef} in radiology report generation, and excels in visual localization, outperforming the established object detector \citep{Jocher2020-xm}. Additionally, it outperforms specialized solutions in longitudinal studies captioning, region describing, open-ended VQA, and chest pathology classification, with performance validated on 6 external cohorts, proving its robustness and generalizability.

\section{Results}
\begin{table}[t]
    \centering
    \caption{\textbf{Experimental results of 4 vision-language tasks: image captioning (report generation), longitudinal captioning, open-ended visual question answering, and region captioning.} Specifically, the evaluation of radiology reports was conducted on three different sections: findings, impression, and all (concatenation). Results of MAIRA and Med-PaLM M are cited from their papers as their models have not been released. Numbers in brackets are the 95\% confidence intervals. $\downarrow$ indicates that the lower results are better. VQA stands for visual question answering. P-values are calculated between the best and second best RadCliQ scores.}
    \resizebox{\textwidth}{!}{
    \begin{tabular}{l|cccccccc}
    \toprule
    \textbf{Tasks}                     & \textbf{Models}      & \textbf{Eval. section} & \textbf{BLEU-4} & \textbf{BertScore} & \textbf{CheXbert} & \textbf{RadGraph} & \textbf{RadCliQ ($\downarrow$)} & P-value \\ \hline
    \multirow{14}{*}{\makecell{Image Captioning \\ (report generation)}}  & ClsGen              & Findings              & \makecell{11.9 \\\small{[}11.4, 12.3{]}} & 40.5 {[}39.8, 41.1{]} & 42.6 {[}41.9, 43.4{]} & 23.5 {[}22.8, 24.2{]} & 3.28 {[}3.24, 3.33{]} & \multirow{6}{*}{$<$ 0.01}  \\
                                       & BiomedGPT           & Findings              & 12.0 {[}11.5, 12.4{]} & 40.8 {[}40.2, 41.5{]} & 43.3 {[}42.8, 43.7{]} & 23.4 {[}22.7, 24.0{]} & 3.25 {[}3.21, 3.29{]} & \\
                                       & GPT-4               & Findings              & 6.3 {[}5.7, 7.0{]}   & 25.4 {[}24.5, 26.4{]} & 37.9 {[}37.0, 38.7{]} & 11.3 {[}10.7, 12.1{]} & 3.95 {[}3.91, 4.00{]} & \\
                                       & MAIRA             & Findings              & 14.2 {[}13.7, 14.7{]} & -                    & 44.0 {[}43.1, 44.9{]} & 24.3 {[}23.7, 24.8{]} & 3.10 {[}3.07, 3.14{]} & \\
                                       & Med-PaLM M (85B)    & Findings              & 11.5 {[}-,-{]}       & -                    & -                   & 26.7 {[}-,-{]}      & -   &             \\
                                        & \cellcolor{LightCyan}MedVersa            & \cellcolor{LightCyan}Findings              & \cellcolor{LightCyan}17.8 {[}17.2, 18.4{]} & \cellcolor{LightCyan}49.7 {[}49.0, 50.4{]} & \cellcolor{LightCyan}46.4 {[}45.5, 47.4{]} & \cellcolor{LightCyan}28.0 {[}27.3, 28.7{]} & \cellcolor{LightCyan}2.71 {[}2.66, 2.75{]} & \\ \cline{2-9}
                     & ClsGen              & Impression            & 8.5 {[}7.6, 9.3{]}   & 38.0 {[}37.3, 38.6{]} & 48.7 {[}48.0, 49.5{]} & 18.8 {[}18.0, 19.7{]} & 3.25 {[}3.18, 3.33{]} & \multirow{4}{*}{$<$ 0.01} \\
                                       & BiomedGPT           & Impression            & 10.2 {[}9.4, 11.0{]}  & 37.4 {[}36.7, 38.2{]} & 49.2 {[}48.4, 49.9{]} & 20.0 {[}19.2, 20.7{]} & 3.09 {[}3.01, 3.16{]} & \\
                                       & GPT-4               & Impression            & 5.5 {[}4.8, 6.2{]}   & 17.2 {[}16.6, 17.9{]} & 22.5 {[}21.6, 23.4{]} & 6.4 {[}5.7, 7.2{]}   & 5.39 {[}5.31, 5.48{]} & \\
                                       & \cellcolor{LightCyan}MedVersa            & \cellcolor{LightCyan}Impression            & \cellcolor{LightCyan}13.7 {[}12.7, 14.7{]} & \cellcolor{LightCyan}48.9 {[}48.0, 49.8{]} & \cellcolor{LightCyan}52.4 {[}51.3, 53.5{]} & \cellcolor{LightCyan}25.7 {[}24.6, 26.9{]} & \cellcolor{LightCyan}2.66 {[}2.60, 2.71{]} & \\ \cline{2-9}
    & ClsGen    & All & 13.7 [13.0, 14.3] & 42.4 [41.6, 43.1] & 44.3 [43.2, 45.4] & 25.2 [24.4, 26.0] & 3.20 [3.14, 3.25] & \multirow{4}{*}{$<$ 0.01} \\
    & BiomedGPT & All & 14.2 [13.5, 14.8] & 42.0 [41.4, 42.6] & 44.6 [44.1, 45.2] & 25.8 [25.1, 26.5] & 3.07 [3.01, 3.14] & \\
    & GPT-4     & All & 6.7 [6.0, 7.3]    & 19.7 [19.2, 20.3] & 25.5 [24.8, 26.3] & 8.7 [8.1, 9.5]    & 4.99 [4.95, 5.03] & \\
    & \cellcolor{LightCyan}MedVersa  & \cellcolor{LightCyan}All & \cellcolor{LightCyan}16.0 [15.3, 16.7] & \cellcolor{LightCyan}47.4 [46.6, 48.2] & \cellcolor{LightCyan}46.6 [45.3, 47.8] & \cellcolor{LightCyan}30.0 [29.1, 30.8] & \cellcolor{LightCyan}2.74 [2.69, 2.79] & \\\hline
    \multirow{3}{*}{Longitudinal captioning} 
    & EKAID    & All & 40.4 [39.9, 41.0] & 69.1 [68.7, 69.5] & 49.1 [48.7, 49.4] & 20.4 [19.9, 20.9] & 2.19 [2.14, 2.23] & \multirow{3}{*}{$<$ 0.01} \\
    & GPT-4    & All & 9.2 [8.6, 9.7]    & 21.8 [21.2, 22.5] & 27.6 [27.0, 28.3] & 10.3 [9.6, 11.0]  & 4.04 [3.99, 4.10] & \\
    & \cellcolor{LightCyan}MedVersa & \cellcolor{LightCyan}All & \cellcolor{LightCyan}44.7 [43.7, 45.6] & \cellcolor{LightCyan}71.4 [70.6, 72.2] & \cellcolor{LightCyan}50.0 [49.5, 50.6] & \cellcolor{LightCyan}23.7 [22.6, 24.9] & \cellcolor{LightCyan}2.05 [2.01, 2.10] & \\\hline
    \multirow{4}{*}{Open-ended VQA} 
    & PTLM       & All & 25.2 [24.4, 26.0] & 64.7 [64.1, 65.5] & 78.3 [77.3, 79.2] & 30.4 [29.7, 31.0] & 1.64 [1.57, 1.71] & \multirow{4}{*}{$<$ 0.01}\\
    & BiomedGPT  & All & 27.1 [26.5, 27.7] & 67.2 [66.5, 67.9] & 80.6 [79.8, 81.3] & 31.6 [31.0, 32.1] & 1.45 [1.37, 1.52] &\\
    & GPT-4      & All & 11.3 [10.9, 11.8] & 31.3 [30.8, 31.9] & 41.5 [40.9, 42.1] & 19.8 [19.3, 20.4] & 3.51 [3.44, 3.57] &\\
    & \cellcolor{LightCyan}MedVersa   & \cellcolor{LightCyan}All & \cellcolor{LightCyan}31.2 [30.7, 31.8] & \cellcolor{LightCyan}76.5 [75.9, 77.1] & \cellcolor{LightCyan}85.1 [84.6, 85.6] & \cellcolor{LightCyan}33.4 [32.7, 34.2] & \cellcolor{LightCyan}1.09 [1.06, 1.12] &\\ \hline
    \multirow{4}{*}{Region captioning} 
    & MiniGPT & All & 5.1 [4.6, 5.5]    & 36.6 [36.3, 37.0] & 55.3 [54.9, 55.8] & 18.3 [17.9, 18.6] & 3.08 [3.05, 3.13] &  \multirow{4}{*}{$<$ 0.01}\\
    & BiomedGPT  & All & 4.5 [4.1, 5.0]    & 32.3 [32.9, 32.6] & 48.4 [48.0, 48.9] & 13.4 [13.1, 13.7] & 3.62 [3.57, 3.67] & \\
    & GPT-4      & All & 3.2 [2.9, 3.5]    & 21.5 [21.0, 22.1] & 19.8 [19.2, 20.4] & 5.6 [5.1, 6.2]    & 5.38 [5.33, 5.42] & \\
    & \cellcolor{LightCyan}MedVersa   & \cellcolor{LightCyan}All & \cellcolor{LightCyan}8.4 [8.2, 8.7]    & \cellcolor{LightCyan}43.8 [43.6, 44.1] & \cellcolor{LightCyan}60.7 [60.4, 61.1] & \cellcolor{LightCyan}22.8 [22.5, 23.1] & \cellcolor{LightCyan}2.70 [2.68, 2.71] & \\
    \bottomrule                        
    \end{tabular}}
    \label{tab:1}
\end{table}

\subsection{Enhancing medical image understanding with generalist learning}
Generalist learning across all tasks outperforms training with task-specific data, as demonstrated by Fig. \ref{fig:3}{\color{Magenta}a}, highlighting the performance gains achieved through the integration of generalist learning, particularly in the context of medical image interpretation. Chest radiographs were chosen as the representative imaging modality due to their inherent complexity and the wide variety of analytical tasks they encompass. This modality provides a multifaceted challenge, making it an ideal platform for assessing the robustness and versatility of the generalist learning paradigm.

The results show that generalist learning led to a mean performance improvement of 6.4\% compared to models trained exclusively on vision-language data. This indicates that vision-centric training within the generalist learning framework enhances the model's ability to process visual information comprehensively, enabling better interpretation and utilization of visual cues across diverse contexts. Additionally, combining vision-centric and vision-language tasks during generalist learning resulted in a mean performance improvement of 5.3\% over models trained solely on vision-centric tasks. This finding highlights the importance of linguistic supervision in improving the model's capacity for visual comprehension and reasoning. By integrating both language and structured supervision into the training process, MedVersa develops more generalized and comprehensive representations, capturing a broader range of features, relationships, and semantic meanings. These results underscore the significance of developing generalist medical AI models capable of supporting multimodal outputs and effectively learning from multimodal supervision to achieve optimal performance and generalization across a wide spectrum of medical image interpretation tasks.

\subsection{Report generation}
Table \ref{tab:1} presents the evaluation results on three sections: findings, impression, and all (concatenation of findings and impression) using five evaluation metrics: BLEU-4 \citep{Papineni2002-lt}, BertScore \citep{Zhang2019-kn}, CheXbert \citep{Smit2020-dv}, RadGraph \citep{Jain2021-bt}, and RadCliQ \citep{Yu2023-nq}. For the findings section, among the baselines, MAIRA \citep{Hyland2023-al} achieves a higher BLEU-4 score of 14.2, while Med-PaLM M \citep{Tu2023-ef} produces a better RadGraph score of 26.7, both of which are the current state-of-the-art. Since MAIRA and Med-PaLM M are not publicly accessible, we additionally included two extra baselines, ClsGen \citep{Nguyen2021-gq} and BiomedGPT \citep{Zhang2024-ec}, for consistent comparisons across different sections. Both ClsGen and BiomedGPT achieve higher BLEU-4 scores than Med-PaLM M.

\begin{table}[t]
\centering
\caption{\textbf{External validation results.} We evaluated six capabilities (i.e., image captioning, image classification, object detection, 2D image segmentation, open-ended visual question answering, and region captioning) on unseen external cohorts. For each capability, we compared MedVersa with its best performing baseline from Table 1. For classification, detection, and segmentation tasks, we used the mean F1 score, mean IoU (Intersection over Union), and mean DICE score as the evaluation metrics, respectively. For other tasks, we reported the results of RadCliQ. Numbers in brackets are the 95\% confidence intervals. P-values are presented. $\downarrow$ indicates that the lower results are better.}
\resizebox{\textwidth}{!}{
\begin{tabular}{lllllll}
\toprule
\textbf{Capabilities} & \textbf{Datasets} & \textbf{Models} & \textbf{Metrics} & \textbf{Results} & \textbf{95\% CIs} & \textbf{P-values} \\
\midrule
\multirow{2}{*}{Image captioning} 
& \multirow{2}{*}{IUX-ray} & BiomedGPT & \multirow{2}{*}{RadCliQ ($\downarrow$)} & 3.12 & [3.04, 3.17] & \multirow{2}{*}{P $<$ 0.01} \\
& & \cellcolor{LightCyan}MedVersa & & \cellcolor{LightCyan}2.57 & \cellcolor{LightCyan}[2.54, 2.60] & \\
\midrule
\multirow{2}{*}{Image classification} 
& \multirow{2}{*}{CheXpert} & DAM & \multirow{2}{*}{Mean F1 score} & 0.653 & [0.633, 0.669] & \multirow{2}{*}{P $<$ 0.01} \\
& & \cellcolor{LightCyan}MedVersa & & \cellcolor{LightCyan}0.734 & \cellcolor{LightCyan}[0.712, 0.756] & \\
\midrule
\multirow{2}{*}{Object detection} 
& \multirow{2}{*}{NIH ChestX-ray} & YOLO & \multirow{2}{*}{Mean IoU} & 0.223 & [0.210, 0.235] & \multirow{2}{*}{P $=$ 0.082} \\
& & \cellcolor{LightCyan}MedVersa & & \cellcolor{LightCyan}0.239 & \cellcolor{LightCyan}[0.225, 0.254] & \\
\midrule
\multirow{2}{*}{2D image segmentation} 
& \multirow{2}{*}{CheXmask} & nnSAM & \multirow{2}{*}{Mean DICE score} & 0.923 & [0.917, 0.928] & \multirow{2}{*}{P $<$ 0.01} \\
& & \cellcolor{LightCyan}MedVersa & & \cellcolor{LightCyan}0.955 & \cellcolor{LightCyan}[0.952, 0.957] & \\
\midrule
\multirow{2}{*}{Open-ended VQA} 
& \multirow{2}{*}{IUX-ray} & BiomedGPT & \multirow{2}{*}{RadCliQ ($\downarrow$)} & 1.68 & [1.62, 1.76] & \multirow{2}{*}{P $<$ 0.01} \\
& & \cellcolor{LightCyan}MedVersa & & \cellcolor{LightCyan}1.12 & \cellcolor{LightCyan}[1.07, 1.17] & \\
\midrule
\multirow{2}{*}{Region captioning} 
& \multirow{2}{*}{MS-CXR} & MiniGPT & \multirow{2}{*}{RadCliQ ($\downarrow$)} & 3.43 & [3.38, 3.48] & \multirow{2}{*}{P $<$ 0.01} \\
& & \cellcolor{LightCyan}MedVersa & & \cellcolor{LightCyan}3.29 & \cellcolor{LightCyan}[3.23, 3.35] & \\
\bottomrule
\end{tabular}}
\label{tab:2}
\end{table}

The proposed MedVersa was evaluated across all sections and metrics. It outperforms all baselines in the findings section with a BLEU-4 score of 17.8 (vs. 14.2 of MAIRA), a CheXBert score of 46.4 (vs. 44.0 of MAIRA), and a RadGraph score of 28.0 (vs. 26.7 of Med-PaLM M), establishing its superiority and setting the new state-of-the-art in capturing both the linguistic and clinical aspects of radiology reporting. Particularly noteworthy is that the results of Med-PaLM M were obtained from a significantly larger model, with ten times more parameters than those of MedVersa. This implies that the latter model is more advantageous in terms of training and inference efficiency. For the impression section, MedVersa surpasses BiomedGPT in all evaluation metrics, and the same superiority is maintained when all sections are combined. We performed external validation on the IUX-ray dataset \citep{Demner-Fushman2016-kk} (Table \ref{tab:2}), where MedVersa keeps maintaining a notable advantage over BiomedGPT.

\subsection{Clinician evaluation}
\begin{figure}[!htp]
    \centering
    \includegraphics[width=\textwidth]{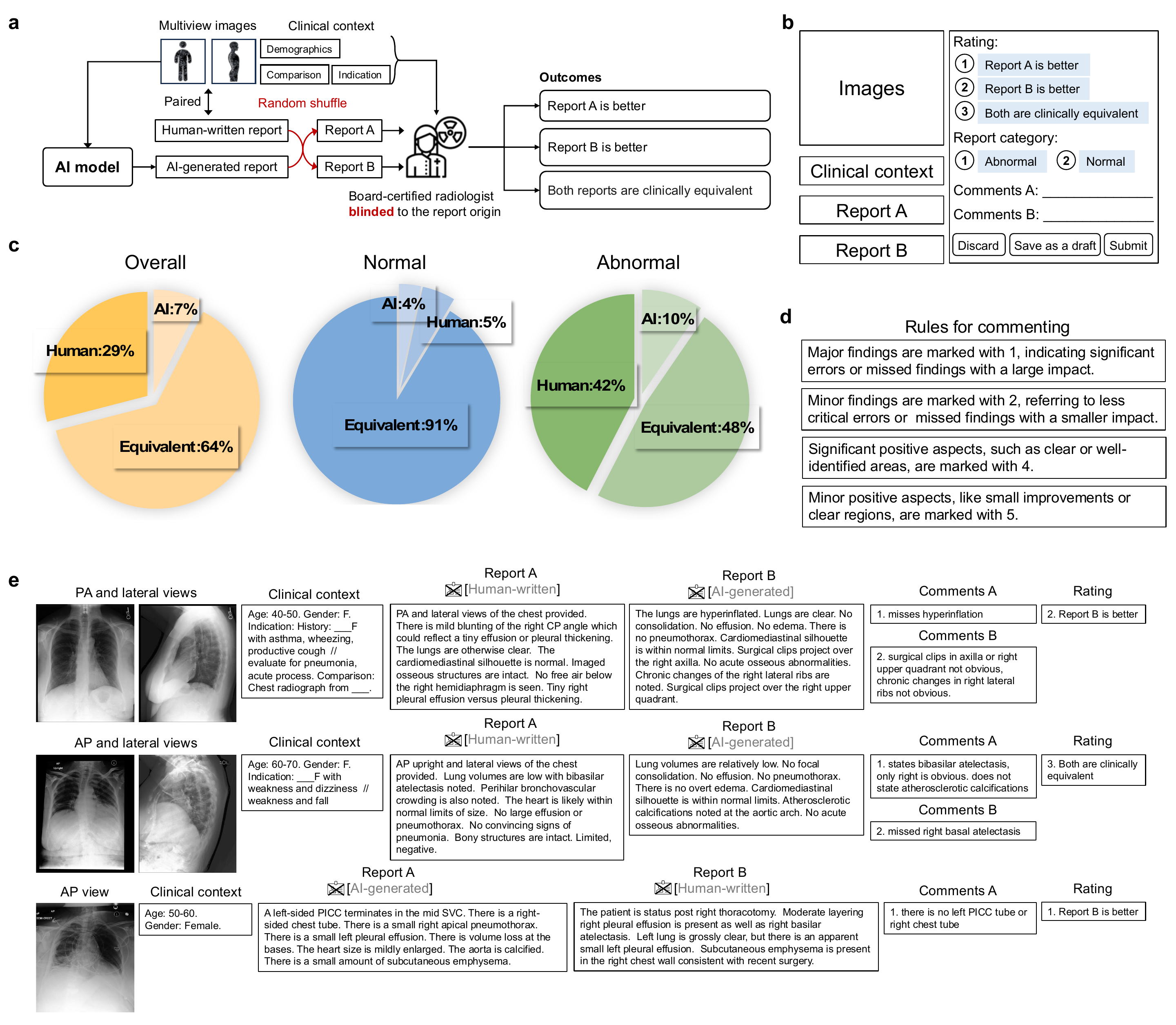}
    \caption{\textbf{Clinical evaluation.} \textbf{a,} Evaluation pipeline showing the blinded assessment process where board-certified radiologists compare randomly shuffled, deidentified human-written and AI-generated reports based on multiview images and clinical context. \textbf{b,} Interface design for radiologist evaluation, including rating options and report categorization. \textbf{c,} Quantitative assessment results showing the distribution of preferences for human versus AI reports across overall, abnormal, and normal cases. \textbf{d,} Protocols for making comments on reports. \textbf{e,} Cases demonstrating the comparison between human-written and AI-generated reports, including image, clinical context, report content, radiologist comments, and ratings.}
    \label{fig:4}
\end{figure}

The evaluation pipeline was designed to ensure objective and unbiased assessment of radiological reports (Fig. \ref{fig:4}{\color{Magenta}a}). Board-certified radiologists received medical images along with essential clinical information, including patient demographics, comparative data, and study indications. This clinical context helped guide their interpretations while reflecting real-world scenarios, ensuring that evaluations remained grounded in practical clinical applications. The same imaging and clinical data were processed through the AI model to generate automated reports, maintaining consistency in the input data across both human and AI assessments. To maintain objectivity, the human-written and AI-generated reports underwent a random shuffling process, becoming ``Report A'' and ``Report B''. This blinding mechanism prevented radiologists from identifying report origins or being influenced by recognizable patterns, thereby eliminating potential biases that could skew the evaluation results. Fig. \ref{fig:4}{\color{Magenta}b} presented the interface design that radiologists used during the evaluation process, featuring an intuitive layout that facilitated efficient and systematic review of the reports. During each evaluation round, radiologists examined two reports alongside the corresponding image and clinical data, choosing whether Report A was superior, Report B was superior, or both reports were clinically equivalent. This three-choice system allowed for nuanced comparison while capturing the practical reality that reports might be equally suitable for clinical use.

The comparative analysis revealed that AI-generated reports demonstrated remarkable performance across various study types (Fig. \ref{fig:4}{\color{Magenta}c}). In the overall assessment, where radiologists remained blinded to report origins, AI reports matched or exceeded human-generated reports in 71\% of cases. This high percentage of equivalence or superiority suggested that AI systems had reached a significant milestone in medical report generation. For abnormal studies with more complex findings, AI maintained strong performance, with reports being equivalent to or preferred over human reports in 58\% of cases. This performance in challenging cases demonstrated the AI's capability to handle intricate medical interpretations that traditionally required substantial human expertise. The AI system particularly excelled in normal studies, where 95\% of its reports were either equivalent to or favored over human reports. This exceptional performance in routine cases suggested a potential role for AI in streamlining workflow for standard examinations, allowing radiologists to focus more attention on complex cases.

A structured scoring system was implemented to ensure consistent evaluation of report quality (Fig. \ref{fig:4}{\color{Magenta}d}). Radiologists employed a numbered scoring system where 1 signified major findings with significant errors, 2 indicated minor findings and less critical errors, 4 represented significant positive aspects such as well-identified areas, and 5 denoted minor positive aspects including small improvements or clear regions. This systematic approach to scoring enabled quantitative analysis of report quality while capturing both critical errors and positive attributes. The evaluation process culminated in detailed case comparisons (Fig. \ref{fig:4}{\color{Magenta}e}), which presented examples featuring medical images, patient context, report content, expert commentary, and final ratings for both human-written and AI-generated reports.

\subsection{User study for MedVersa-assisted report writing}
\begin{figure}[htbp]
    \centering
    \includegraphics[width=\textwidth]{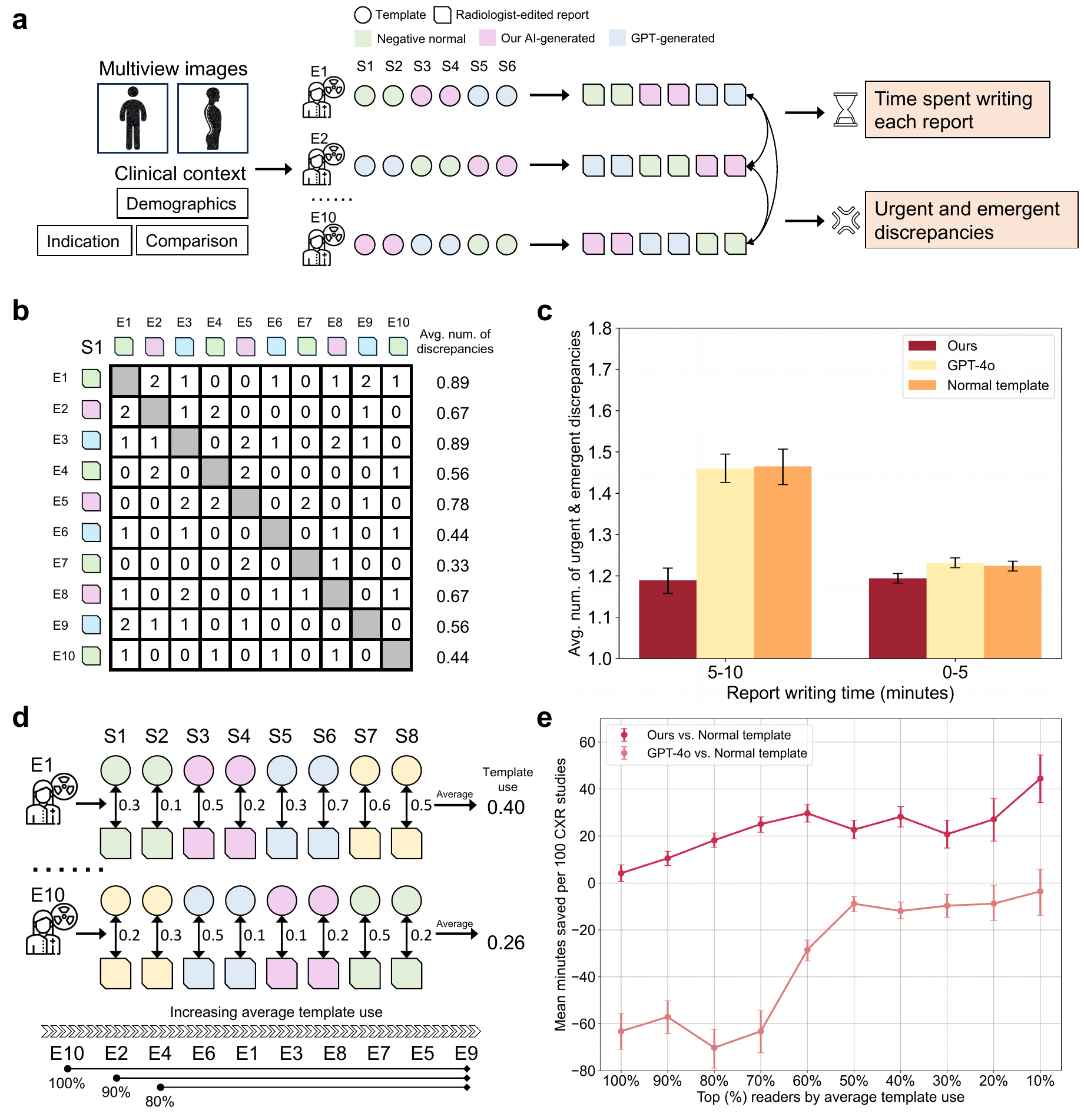}
    \caption{\textbf{Clinical impact study.} \textbf{a,} Study design showing multiple board-certified radiologists (E1-E10) editing reports based on different templates (negative normal, our AI-generated, and GPT-generated) for multiple studies (S1-S6), measuring time spent and discrepancies. \textbf{b,} Matrix visualization of urgent and emergent discrepancies across experts and studies, with average number of discrepancies per expert. \textbf{c,} Comparison of average number of urgent and emergent discrepancies across different report writing time intervals for different template types. \textbf{d,} Detailed template usage patterns across different studies and experts, with average template utilization rates shown. \textbf{e,} Time saved in report writing compared to normal template usage, stratified by top percentage of readers based on average template use.}
    \label{fig:5}
\end{figure}
Our study established a comprehensive workflow for evaluating the clinical impact of AI-generated radiology reports through collaboration between radiologists and AI, as illustrated in Fig. \ref{fig:5}{\color{Magenta}a}. We recruited ten board-certified radiologists through a specialized medical image annotation company in India. The study utilized randomly selected chest radiographs from the MIMIC-CXR dataset. We developed a custom evaluation platform (Fig. \ref{fig:5}{\color{Magenta}b}) that integrated several key components: a DICOM viewer for image interpretation, a worklist interface, and a report editing interface where radiologists could modify and finalize their reports. The platform's integrated timer tracked time spent on each case. The preparation phase included providing participants with a detailed study task protocol and an instructional video, completing a standardized training phase of 25 cases to ensure platform familiarity, and conducting an interactive orientation meeting where we demonstrated platform functionality, reviewed task requirements, and addressed any remaining questions about the study procedures.

For the main evaluation, each radiologist interpreted 75 unique chest radiographs under three different reporting scenarios: (1) starting with a standard negative template (as per routine radiology workflow), (2) starting with a GPT-4o-generated report draft, or (3) starting with our AI-generated report draft. Cases were randomly and evenly distributed across these three scenarios, with each radiologist reading 25 cases per scenario. For each case, radiologists were tasked with reviewing the chest radiograph and modifying the provided template or draft to produce a clinically accurate final report that would meet the standards of clinical practice. To eliminate potential recall bias and need for washout periods, each radiologist interpreted a case only once, while the same case was interpreted under different scenarios across different radiologists, ensuring balanced distribution of case complexity. We evaluated two key dimensions of reporting performance: discrepancy and efficiency. For discrepancy, we compared the ``findings'' section of each final report against those of the other nine radiologists for the same case, using an adapted version of FineRadScore \citep{Huang2024-vi}. For each report, we calculated its FineRadScore against each of the other nine reports and averaged these scores to obtain an average discrepancy score (Fig. \ref{fig:5}{\color{Magenta}c}), effectively penalizing reports that deviated from the consensus. Specifically, we focused on urgent and emergent discrepancies, as these most directly impact patient care, highlighting the critical importance of achieving consensus in high-stakes scenarios. We chose this peer-to-peer comparison over using the original ground truth reports because those reports often contain information not available to our readers or models—such as comparisons to prior imaging studies—which could introduce bias. By focusing on the radiologists’ reports, all based on the same available data, we ensured that our assessment accurately reflected consistency within the group. Reporting efficiency was measured through the platform's integrated timer.

The analysis of urgent and emergent discrepancies in radiological reports reveals significant variations in performance across different AI systems and time intervals. As shown in Fig. \ref{fig:5}{\color{Magenta}d}, we examine these discrepancies across two time intervals (5–10 and $<$5 minutes) for three different scenarios: our model, GPT-4o, and the normal template. Our model consistently achieves lower discrepancy rates across different writing time periods, indicating its effectiveness in managing discrepancies across varying writing durations. Specifically, in the 5–10 minute interval, our model demonstrates a notable advantage over GPT-4o and the normal template, achieving the lowest average number of urgent/emergent discrepancies. Our model maintains approximately 20\% fewer discrepancies compared to the other systems, reflecting greater consistency and reliability in identifying urgent and emergent issues during report writing. GPT-4o and the normal template perform similarly during this interval, with neither showing a clear advantage in reducing discrepancies over time..

The impact of AI assistance on radiology reporting time varies significantly based on radiologists' template usage patterns. Template use is measured by calculating the average semantic similarity between the template referred (our model, GPT-4o, or a standard negative template) and the corresponding final report (Fig. \ref{fig:5}{\color{Magenta}e}) using the RaTEScore metric \citep{Zhao2024-vj}. A higher similarity score indicates a stronger preference for using the template, which may be because the template meets their requirements for accuracy and/or aligns with their personal reporting style. In practice, this means that radiologists with high template use often make focused edits to ensure clinical accuracy while retaining much of the original template, whereas radiologists with low template use tend to discard templates or large parts of them entirely and rewrite the report from scratch. As shown in Fig. \ref{fig:5}{\color{Magenta}f}, our model consistently reduces reporting time across radiologist groups, with efficiency gains becoming more pronounced as reliance on templates increases. This upward trend highlights our model's adaptability, with time savings growing steadily and peaking at around 50 minutes saved per 100 studies. Our hypothesis is that the extent to which radiologists use templates may influence how AI impacts their work. In contrast, GPT-4o shows less favorable results, as demonstrated in Fig. \ref{fig:5}{\color{Magenta}f}. For radiologists with low template use, GPT-4o leads to a notable increase in time required, with the mean time cost rising by approximately 60 minutes. Even as template use increases, GPT-4o only begins to show marginal time savings in the top 20\% of radiologists, and its performance remains far below that of our model at all levels of template reliance. The comparison highlights our model's ability to cater to a wide spectrum of radiologist preferences, offering consistent and increasing time savings as template reliance grows, whereas GPT-4o struggles to achieve similar levels of efficiency, particularly for radiologists less inclined to use templates.

\subsection{Vision-centric tasks}
\begin{figure}[!htp]
    \centering
    \includegraphics[width=\textwidth]{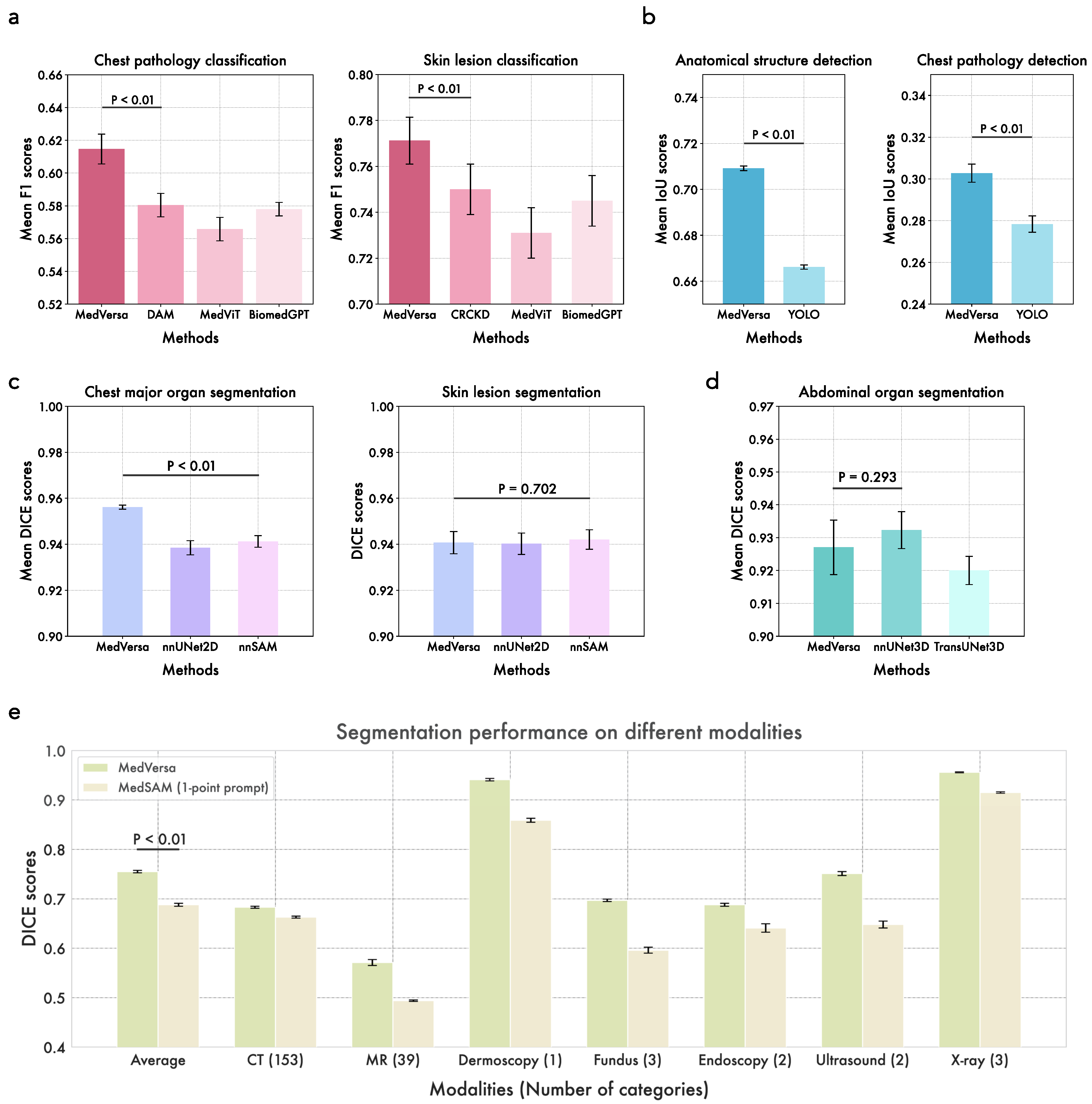}
    \caption{\textbf{Experimental results of vision-centric tasks.} \textbf{a,} MedVersa was compared to three baseline models: DAM (Deep AUC Maximization), MedViT, and BiomedGPT on  chest pathology classification. For skin lesion classification, we replaced DAM with CRCKD (Categorical Relation-preserving Contrastive Knowledge Distillation), a model specifically designed for this task. \textbf{b,} We compared MedVersa against YOLO (version five) on two detection tasks: anatomical structure and chest pathology detection. \textbf{c,} For 2D image segmentation, we primarily compared MedVersa with nnUNet2D and nnSAM for segmenting major organs in the chest and skin lesions. \textbf{d,} For 3D image segmentation, MedVersa was compared with nnUNet3D and TransUNet3D for segmenting abdominal organs. \textbf{e,}  We investigated the development of a versatile segmentation model for various imaging modalities, extending the functionality of MedVersa. The baseline approach, MedSAM, is a segment anything model finetuned specifically on medical segmentation data. The evaluation metrics of classification, detection, and segmentation tasks are F1, IoU (Intersection over Union), and DICE similarity scores, respectively.}
    \label{fig:6}
\end{figure}

Fig. \ref{fig:6}{\color{Magenta}a} presents the experiment results of image classification. For chest pathology classification, MedVersa demonstrates superior performance over DAM \citep{Yuan2021-zj}, MedViT \citep{Manzari2023-sj}, and BiomedGPT \citep{Zhang2024-ec} with an average F1 score of 0.615. This pattern of outperformance extends in 29 out 33 pathologies (see Fig. \ref{fig:10}{\color{Magenta}a}), including both common (e.g., lung opacity, pulmonary edema, spinal fracture) and less common ones (e.g., hydropneumothorax, bronchiectasis), which indicates MedVersa's strong diagnostic accuracy across various conditions. In skin lesion classification, the advantage of MedVersa is also noticeable. The average F1 score of MedVersa is 0.772, appreciably above the scores of CRCKD \citep{Xing2021-xp}, MedViT, and BiomedGPT, underscoring MedVersa's effectiveness in classifying skin conditions (Fig. \ref{fig:6}{\color{Magenta}a}). It is worth noting that MedVersa outperforms CRCKD by significant margins in benign keratosis-like lesions (bkl), which has a diverse range of subtypes (see Fig. \ref{fig:10}{\color{Magenta}b}). This further demonstrates the generalization ability of MedVersa. For external validation (Table \ref{tab:2}), MedVersa again surpasses DAM by a large margin on CheXpert \citep{Irvin2019-hd}, which also exceeds the mean performance of radiologists (F1 score: 0.734 vs. 0.610) \citep{Tu2023-ef}.

\begin{table}[t]
\centering
\caption{\textbf{Segmentation results of skin lesions and abdominal organs.} For abdominal organ segmentation, the red, blue, green, and aqua colors represent the pancreas, liver, kidney, and spleen, respectively.}
\resizebox{\textwidth}{!}{
\begin{tabular}{c|c|c|c}
\toprule
\multicolumn{2}{c|}{Skin leison segmentation} & \multicolumn{2}{c}{Abdominal organ segmentation} \\ \hline
\textbf{Ground truth} & \textbf{MedVersa} & \textbf{Ground truth} & \textbf{MedVersa} \\ \hline
\includegraphics[width=0.2\textwidth]{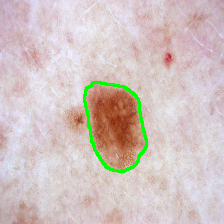} & 
\includegraphics[width=0.2\textwidth]{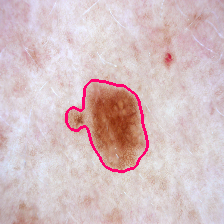} & 
\reflectbox{\includegraphics[width=0.2\textwidth, angle=90, origin=c]{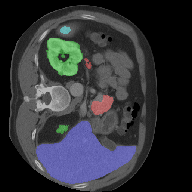}} & 
\reflectbox{\includegraphics[width=0.2\textwidth, angle=90, origin=c]{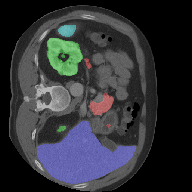}} \\ 

\includegraphics[width=0.2\textwidth]{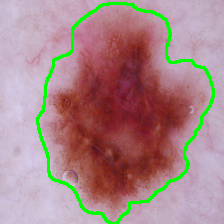} & 
\includegraphics[width=0.2\textwidth]{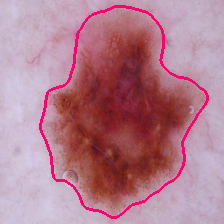} & 
\reflectbox{\includegraphics[width=0.2\textwidth, angle=90, origin=c]{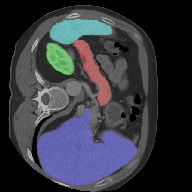}} & 
\reflectbox{\includegraphics[width=0.2\textwidth, angle=90, origin=c]{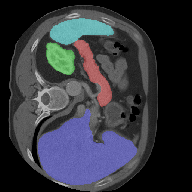}} \\ 


\includegraphics[width=0.2\textwidth]{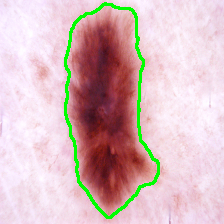} & 
\includegraphics[width=0.2\textwidth]{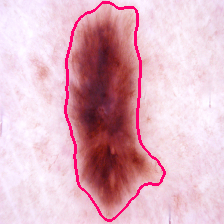} & 
\reflectbox{\includegraphics[width=0.2\textwidth, angle=90, origin=c]{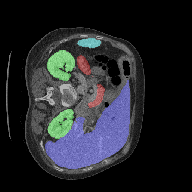}} & 
\reflectbox{\includegraphics[width=0.2\textwidth, angle=90, origin=c]{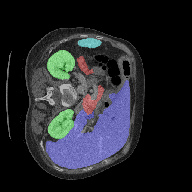}} \\ \bottomrule
\end{tabular}}
\label{tab:3}
\end{table}

For object detection, MedVersa exhibits competitive performance, surpassing YOLO \citep{Jocher2020-xm} by noticeable, consistent margins in the detection of a variety of anatomical structures (Fig. \ref{fig:5}{\color{Magenta}b}), with most IoU scores on certain structures surpassing 0.6 (Fig. \ref{fig:11}{\color{Magenta}a}). It shows particularly high effectiveness in the detection of lung zones. When identifying chest pathologies (Fig. \ref{fig:5}{\color{Magenta}b}),  MedVersa maintains a higher average performance compared to YOLO (0.303 vs. 0.278), and the superiority is also notable in the detection of 27 out of 33 conditions (Fig. \ref{fig:11}{\color{Magenta}b}). On the external cohort NIH ChestXray, MedVersa also outperforms YOLO by an average of nearly two percent in detecting common chest pathologies (Table \ref{tab:2}).

Regarding segmentation tasks, MedVersa demonstrates competitive results, performing competitively to nnUNet \citep{Isensee2021-fu} and nnSAM \citep{Li2023-at}. All three approaches perform fairly well in segmenting major chest organs and skin lesions (Fig. \ref{fig:5}{\color{Magenta}c} and Fig. \ref{fig:12}{\color{Magenta}a}). Nonetheless, MedVersa outperforms nnUNet and nnSAM by significant margins in chest major organ segmentation. In the task of abdominal organ segmentation (Fig. \ref{fig:5}{\color{Magenta}d} and Fig. \ref{fig:12}{\color{Magenta}b}), MedVersa also shows competitive performance to nnUNet3D which uses more complex and time-consuming data augmentation techniques. Table \ref{tab:3} showcases the segmentation results of skin lesions and abdominal organs.
    
Fig. \ref{fig:5}{\color{Magenta}e} presents a comparative analysis of the segmentation performance between MedVersa and MedSAM (1-point prompt) models across varied imaging modalities. MedVersa, based on its architecture, shows a consistently higher DICE score compared to MedSAM in different modalities, indicating its superior ability to accurately segment different types of medical images. The modalities examined include CT, MR, dermoscopy, fundus, endoscopy, ultrasound, and chest x-ray. On average, MedVersa achieves a DICE score of 0.755, significantly outperforming the baseline approach  which is a SAM model \citep{Kirillov2023-ia} finetuned on the medical segmentation data. We believe that this performance advantage can be partly attributed to the features produced by the language model which may encode underlying relationships between different categories.

\subsection{Longitudinal captioning, visual question answering, and region captioning}
In longitudinal captioning, the model is typically tasked with drawing a comparative conclusion between two groups of images collected at different periods. This presents a significant challenge for image interpretation, as models must work with multiple images to analyze various anatomical structures, extracting features and identifying subtle disease-related changes. As shown in Table \ref{tab:1}, the baseline method EKAID builds complex anatomical structure-aware graphs to encode anatomical and disease features for recognizing the differences between CXR studies. In contrast, MedVersa adopts a straightforward yet effective way to process longitudinal images (see. Fig. 7e). Moreover, MedVersa largely outperforms EKAID across different metrics (BLEU-4: 44.7 vs. 40.4, BertScore: 71.4 vs. 69.1, CheXbert: 50.0 vs. 49.1, RadGraph: 23.7 vs. 20.4, RadCliQ: 2.05 vs. 2.19). The same performance advantage is also present in the comparison with GPT-4.

As Table \ref{tab:1} displays, MedVersa outperforms PTLM \citep{Van_Sonsbeek2023-ni}, a leading model for open-ended medical VQA, by an average of six percent across BLEU-4, BertScore, CheXbert, and RadGraph metrics. Additionally, MedVersa achieves a 30\% lower RadCliQ score compared to PTLM. The 95\% confidence intervals suggest that MedVersa's improvements are statistically significant. Results on the external cohort further confirm MedVersa’s advantages (Table \ref{tab:2}). Notably, MedVersa also outperforms BiomedGPT and GPT-4 by substantial margins. To enhance MedVersa’s medical VQA capabilities, we also focused on its ability to process multimodal images. By fine-tuning the pretrained 2D vision encoder and adapter while keeping the language model frozen, we trained the model using the PMC-VQA dataset from PubMed Central Open Access (which includes multimodal medical images) and PathVQA (histopathology images), following the BiomedGPT training approach. Validation on the PathVQA and VQA-RAD test sets highlighted MedVersa’s superior performance. On PathVQA, MedVersa achieved superior accuracy of 64.6\%, compared to BiomedGPT’s 58.0\%. This included 46.5\% accuracy on open-ended questions (versus BiomedGPT’s 28.0\%) and 82.8\% on closed-ended questions (versus BiomedGPT’s 88.0\%). In zero-shot classification on VQA-RAD, MedVersa reached a mean accuracy of 63.3\% ± 2.5\%, outperforming BiomedGPT (54.7\% ± 5.7\%) and GPT-4 (53.0\% ± 6.7\%).
    
In the region captioning task, MedVersa demonstrates a clear advantage over MiniGPT across multiple metrics, even though MiniGPT can outperform GPT-4 and BiomedGPT. The RadCliQ score, which comprehensively evaluates the lexical and clinical significance of generated text,  is substantially lower for MedVersa at 2.70 versus 3.08 for MiniGPT, suggesting captions of MedVersa are semantically more aligned with reference standards. The result from an external cohort further confirms the benefit of MedVersa, as shown in Table \ref{tab:2}.

\section{Discussion}
To the best of our knowledge, MedVersa is the first GMAI model that supports multimodal outputs, inputs, and on-the-fly task specification. Trained on MedInterp, a medical dataset encompassing 11 different tasks across seven imaging modalities, MedVersa sets the new state-of-the-art in report generation and outperforms highly competitive specialist models in both vision-language and vision-centric tasks. The development of MedVersa potentially unlocks new opportunities to build more versatile GMAI models. More detailed perspectives are provided in the following.

\textbf{MedVersa integrates visual and linguistic supervision through its multimodal-output design.} MedVersa distinguishes itself from previous endeavors by  seamlessly incorporating both visual and textual guidance in its training process. This unique approach allows MedVersa to tackle a wide range of medical tasks, from generating radiology reports to segmenting medical images. The model's ability to assimilate knowledge from various input types and generate multimodal outputs results in the development of general  and robust shared representations, which helps boost the model accuracy on the tasks and alleviate potential biases in the data. The incorporation of multimodal outputs in MedVersa's also aligns with the latest progress in generative AI, where the use of varied and all-encompassing training data has yielded promising results. By gaining insights from both visual and textual cues, MedVersa constructs a more comprehensive grasp of medical information, paving the way for more precise and dependable diagnoses. Its capacity to adapt to impromptu task specifications renders MedVersa a multifaceted and flexible instrument for diverse clinical applications, establishing its place as a useful resource in medical AI for thorough diagnostics.

\textbf{Large language models act as optimizable orchestrators.} Unlike previous endeavors that used large language models as standalone language predictors, the large language model in MedVersa transcends its traditional role by acting as an optimizable orchestrator capable of interpreting medical vision-language data and coordinating with vision modules. This design allows MedVersa to leverage the strengths of both the large language model and specialist components, resulting in a more comprehensive and effective system for medical image interpretation. The integration of vision modules within MedVersa  enhances its capability in areas where language-based models traditionally falter, such as detailed image analysis required in chest abnormality detection and skin lesion segmentation. The comprehensive approach, combining the contextual decision-making of the large language model with the precision of vision modules, offers a more robust and versatile diagnostic tool. This new orchestration represents a new step beyond the limitations of previous medical foundation models, offering a new perspective of integrating large language models into generative multimodal medical AI.

\textbf{Impact of dataset composition.} The composition of our dataset significantly influences the scope and generalizability of our findings. While we have achieved comprehensive coverage for chest X-rays across various task types, including classification, detection, and report generation, the diversity of tasks for other imaging modalities is currently limited. Specifically, for modalities such as CT, MRI, endoscopy, dermoscopy, ultrasound, and fundus imaging, our annotations are predominantly segmentation masks. This imbalance in task representation across modalities potentially constrains the model’s ability to generalize across a broader range of medical imaging tasks. To address this limitation, our future work will focus on expanding the diversity of tasks for non-chest X-ray modalities. We plan to incorporate a wider range of annotation types, such as classification labels, bounding boxes, and descriptive reports, for these other imaging modalities. Additionally, we recognize the importance of incorporating more text-only data to enhance the model's language understanding capabilities in medical contexts. These planned expansions aim to create a more balanced and comprehensive dataset, which should lead to more robust and versatile AI models for medical image interpretation across various specialties and task types.

\textbf{Balancing performance across modalities and tasks.} A key challenge in developing generalist medical AI models lies in maintaining consistent performance across different imaging modalities and tasks, given the inherent data imbalance in medical datasets. Our experiments demonstrate that MedVersa addresses this challenge through its domain-aware minibatch gradient descent approach. When extending to new domains like dermoscopy and CT volumes, the model maintained its chest X-ray interpretation capabilities. This stability stems from our localized gradient update mechanism, where task-specific minibatches primarily influence network parameters relevant to the current task-modality pair. The approach enables the model to develop shared representations that support transfer between tasks rather than interference. However, we recognize limitations in our current implementation, as random task selection within each modality may not fully account for differences in task complexity. Future work could explore adaptive sampling strategies that adjust based on model performance and task difficulty, while incorporating continual learning techniques could enhance the model's ability to maintain balanced performance across an expanding set of medical imaging tasks and modalities.

\textbf{Extensible GMAI and beyond.} The system design features a notable level of extensibility, allowing for the practical integration of new vision modules into its existing framework. This aspect of MedVersa enables it to adapt and grow in response to evolving medical imaging techniques and diagnostic requirements. Differing from traditional medical AI models, MedVersa integrates the large language model in the way that provides a extensible platform for the addition of new specialist models as advancements in medical technology occur. This feature ensures that the overall system remains up-to-date and effective in a field characterized by rapid technological changes and emerging diagnostic challenges. As novel medical imaging methods are introduced, MedVersa can be updated to maintain its relevance in the dynamic landscape of medical diagnostics. This modular design not only prepares it for future advancements but also encourages ongoing improvement and innovation within the system. It highlights the potential of MedVersa as an extensible and adaptable solution in medical AI, equipped to address the varied and changing requirements of healthcare practitioners and patients in a continuously evolving medical environment.

\textbf{Potential to streamline clinical workflows.} The impact of our work is prominent in offering a unified solution that can help streamline the clinical workflows with medical AI products. In contrast, task-specific models, designed for individual tasks, may complicate or fragment workflows, necessitating the medical professionals to switch between multiple systems. For instance, in a busy metropolitan hospital, the radiology department faces challenges managing a high volume of diverse imaging tasks daily, from urgent chest X-ray interpretations to CT scans requiring detailed analysis. The introduction of MedVersa allows for a seamless transition between these tasks within a single, integrated platform. Previously, radiologists had to switch between multiple specialist AI modules, each with its own interface and diagnostic focus, leading to inefficiencies and delays in patient care. The comprehensive capability of MedVersa to interpret various types of medical images means that radiologists could efficiently work through their caseloads, significantly reducing the turnaround time for diagnostic reports. This streamlined process not only improves operational efficiency but also ensures that patients receive faster diagnoses, leading to quicker treatment decisions and better outcomes, finally increasing the adoption rate of AI products in real-world clinical settings.

\textbf{AI-Radiologist collaboration.} A critical consideration in understanding the impact of AI on radiologists’ well-being is examining why and how AI may either alleviate or contribute to their challenges. The effectiveness of AI solutions extends beyond their accuracy; factors such as workflow integration, user experience, and the overall ease of the human-AI interaction play pivotal roles. Seamless integration into clinical workflows that enhances rather than disrupts processes, along with user-centered design principles, are essential to achieving meaningful benefits. Importantly, simply accelerating tasks without reducing the cognitive and physical demands on radiology staff may fail to address the underlying contributors to their workload and stress. These nuances underscore a key point: clinical accuracy and efficiency do not automatically translate to clinical utility in healthcare AI. The complexity of human-AI interaction requires a more nuanced approach to assessing AI’s role in clinical practice.

\textbf{Road to full orchestration.} Recent research has delved into the role of LLMs as orchestrators and agents in multiagent systems \citep{Han2024-kh}. Frameworks such as AutoGen \citep{Wu2023-xw} have enabled multiagent conversations, enhancing applications of LLMs. Multiagent debates \citep{Du2023-bu} have been shown to improve factuality and reasoning, while setups that encourage divergent thinking lead to more creative outputs \citep{Liang2023-zx}. Additionally, dynamic LLM-agent networks have been proposed to optimize team performance \citep{Liu2023-ed}, though challenges remain in retaining information and solving complex, multi-step problems \citep{Wang2024-ve}. Emerging patterns in orchestration frameworks highlight evolving strategies for leveraging AI agents \citep{Wang2023-bk}. Achieving full orchestration of MedVersa involves several strategic advancements. First, broadening its dataset scope to encompass a wider array of medical data types, such as detailed electronic health records, comprehensive genetic information, and real-time patient monitoring data, is crucial. This diversification will enhance the diagnostic accuracy of MedVersa by providing a more holistic view of patient health. Second, incorporating cutting-edge AI and machine learning modules, particularly in evolving areas of natural language processing and computer vision, will refine its capability to interpret and analyze complex medical datasets accurately. The development and integration of advanced modules for effective data synthesis and nuanced interpretation are essential for providing comprehensive medical insights. This path also includes rigorous attention to ethical, privacy, and security issues, ensuring MedVersa's operation within a framework that prioritizes patient confidentiality and data integrity. Ultimately, the full orchestration of MedVersa aims to transform healthcare delivery through personalized, efficient, and broad-spectrum medical analyses, leading to superior patient care and optimized healthcare processes.

\textbf{Limitations.} Despite the advancements, there are inherent limitations that warrant consideration. One primary concern lies in the dependency of MedVersa on the quality and diversity of the data used for training the models. If the dataset is not sufficiently varied or representative of the global population, there is a risk of bias in the AI-generated diagnostics, potentially leading to less accurate outcomes for certain demographic groups. Additionally, the complexity of integrating various vision modules with the large language model poses challenges in ensuring seamless interoperability and maintaining the consistency of the system's overall performance. The dynamic nature of MedVersa, while advantageous for adaptability, also raises questions about the long-term manageability and scalability of the system, especially as it continuously evolves to include new modalities and network modules. Moreover, the interpretability and explainability of the decision-making process of MedVersa remains a critical area. The complex interactions between different AI models can obscure the reasoning behind specific diagnostic conclusions, making it challenging for medical professionals to fully understand and trust the recommendations. These limitations underscore the need for ongoing research and development in enhancing the robustness, transparency, and ethical considerations, ensuring it aligns with the highest standards of clinical practice and patient care.

\section{Methods}
\subsection{Datasets and data preprocessing}
We curated MedInterp to train and evaluate medical FMs for multifaceted medical image interpretation. An overview of MedInterp was presented in Table \ref{tab:4}. Specifically, MedInterp consists of 91 publicly available datasets, some of which are associated with more than one task. The details of datasets are listed below, with segmentation-related datasets described in Table \ref{tab:4}. We followed the guideline in SAM-Med to preprocess the segmentation data \citep{Cheng2023-gq}.

\textbf{MIMIC-CXR.} This is a large, publicly accessible dataset comprising 377,110 chest X-rays (CXRs) corresponding to 227,835 radiographic studies performed at the Beth Israel Deaconess Medical Center in Boston, MA \citep{Johnson2019-er,Johnson2019-ug}. The dataset was fully deidentified, and the protected health information was also removed. We referred to the official split and combined studies with ``train'' and ``validate'' tags into the training set, while the rest were included in the test set (for internal validation). The free-text radiology report preprocessing followed the steps in CXR-RePair \citep{Endo2021-fu}. Specifically, we extracted sections of indication, comparison, findings, and impression from free-text radiology reports via keywords matching. Then, we filtered out studies with empty findings and impression sections. After these steps, we can obtain 149,711 (2,144) findings sections and 189,411 (2,212) impression sections. Numbers in parentheses denote the sample size of the test set. Besides, we also extracted complete radiology reports, i.e., reports that have findings and impression sections. This resulted in 122,702 (1,437) complete reports, which were also involved in training and internal validation stages along with sections of findings and impressions. Note that some studies may have more than one CXR, and images of MIMIC-CXR were also used in other tasks.

\textbf{Chest ImaGenome.} This dataset augmented the free-text reports of MIMIC-CXR with local annotations derived from both rule-based natural language processing (NLP) and atlas-based bounding box detection \citep{Wu2021-fp}. These annotations are intricately linked through CXR ontologies developed by radiologists, forming anatomy-centered scene graphs. We followed the data split of MIMIC-CXR to avoid training and test sets leakage. The chest pathology classification task included 235,721 CXRs with annotations of 33 pathologies (Fig. \ref{fig:4}{\color{Magenta}a}). A vast majority of CXRs have bounding box annotations of 36 anatomical structures, leading to 8,425,163 boxes in total. We also exploited the anatomy-centered graph-structured annotation of Chest ImaGenome. For chest pathology detection, we first identified connections between pathologies and anatomies. Next, we can assign bounding boxes of anatomies to associated pathologies that were marked positive. A similar strategy was also adopted for region captioning, where connections between sentences from free-text reports and anatomies were extracted using NLP techniques \citep{Wu2021-fp}. After this, we had textual captions grounded on anatomies. So the task input would be the box coordinates of anatomies, and the output would be the associated captions.

\textbf{Medical-Diff-VQA.} This is a publicly available dataset containing a vast number of question-answer pairs based on CXRs \citep{Hu2023-yf}. To construct this dataset, keywords of abnormality and their attributes were first collected. Then, regular expressions were utilized to detect abnormality/disease keywords within the free-text reports of each patient visit in MIMIC-CXR. These identified keywords served as anchor terms to segment the sentences, and nearby text sections were then scanned for the relevant attribute keywords. The accuracy and completeness of the extracted information have been carefully checked by humans and advanced NLP tools. In practice, we leverage the code in an open source repository to generate the datasets \citep{Hu2023-yf}. Since open-ended visual question answering is our main focus, we reduced the number of yes/no question-answer pairs by setting the \textit{less}\_\textit{yes}\_\textit{no} variable in the code to True. This results in 383,683 normal question-answer pairs, where each pair is associated with one frontal CXR, and 147,269 longitudinal comparisons, where each comparison encompasses two studies, and each study may contain more than one CXR. We used the same data split as in Chest ImaGenome and MIMIC-CXR to avoid training and test information leakage across different datasets. To build a cohort for external validation, we applied the dataset construction code to the free-text reports of IUX-ray \citep{Demner-Fushman2016-kk} to extract 2,883 normal question-answer pairs.

\textbf{PMC-VQA.} The PMC-VQA dataset is a large-scale medical visual question-answering resource that contains 226,946 question-answer pairs associated with 149,075 multimodal medical images \citep{Zhang2023-el}. PMC-VQA was developed to address the limitations of existing datasets in training high-performing generative-based models for medical visual question answering. The dataset's questions vary in complexity, from simple identification tasks to more challenging inquiries requiring specialized knowledge. 

\textbf{PathVQA.} PathVQA is a pioneering medical visual question-answering dataset focused on pathology images, containing 32,795 question-answer pairs derived from 4,998 pathology images \citep{He2020-ox}. The dataset was created by extracting images and captions from pathology textbooks and digital libraries, using a semi-automated pipeline to generate questions that were then manually verified. Unlike previous medical VQA datasets that primarily used radiology images, PathVQA is unique in its focus on pathology and its emphasis on open-ended questions, making it a more challenging and diverse resource for medical AI research. The dataset includes various question types, with 49.8\% being yes/no questions, 40.9\% ``what'' questions, 6.6\% ``where'' questions, and 1.8\% ``how'' questions.

\textbf{VQA-RAD.} VQA-RAD is a pioneering dataset for medical visual question answering, specifically focused on radiology images \citep{Lau2018-ow}. The images are sourced from MedPix, an open-access radiology archive of case reports and teaching cases. It covers various question categories, with the majority being yes/no questions (49.8\%) and ``what'' questions (40.9\%). Note that we only used the test set from VQA-RAD for external validation, which contains 451 question-answer pairs.

\textbf{HAM10000.} This is a bulk collection of multi-source dermatoscopic images of common pigmented skin lesions \citep{Tschandl2018-zi}. The dataset comprises 10,015 image cases and encompasses a diverse collection of significant diagnostic categories within the domain of pigmented lesions. These categories include Actinic keratoses and intraepithelial carcinoma / Bowen's disease (akiec), basal cell carcinoma (bcc), benign keratosis-like lesions (bkl), dermatofibroma (df), melanoma (mel), melanocytic nevi (nv), and vascular lesions (vasc). For the skin lesion classification task, we used 1,511 images from ISIC 2018 task three as the test set for internal validation \citep{Codella2019-nt}. For skin lesion segmentation, we randomly split the dataset into training and test sets. The ratio of the training set to the test set is 9:1. We trained skin classification models on the raw datasets directly without any class balancing skills \citep{Alam2022-hh}.

\textbf{CheXpert.} This is another large public dataset for chest radiograph interpretation, which retrospectively collected the chest radiographic examinations from Stanford Hospital, performed between October 2002 and July 2017 \citep{Irvin2019-hd}. In our case, we used its test set (500 studies, 668 images) with strong ground truth for externally validating the results of the chest pathology classification task. The test set labels were established through the majority vote of annotations from five radiologists, with three of them being the same as those who annotated the validation set, while the other two were randomly selected. Besides, 200 images in the validation set of CheXpert were also used in the task of chest major organ segmentation.

\textbf{IUX-ray.} The dataset contains 7,470 pairs of CXRs and radiology reports \citep{Demner-Fushman2016-kk}. It served as the external validation set for the report generation task. To maintain the consistency of cross-dataset validation, we filtered out reports that do not contain sections of findings and impression simultaneously, resulting in 3,323 studies. Each study has one frontal and one lateral CXRs, associated with one radiology report. Note that images of IUX-ray were also used in the external validation of the open-ended visual question answering task.

\textbf{NIH ChestX-ray.} This dataset includes over 100,000 anonymized CXRs of more than 30,000 individuals from the NIH Clinical Center \citep{Wang2017-sg}. Apart from image-level pathology labels, NIH ChestX-ray also provides a small number of bounding box annotations. In practice, we incorporated the box annotations (577 boxes) of four common chest pathologies - atelectasis, cardiomegaly, effusion, and pneumothorax - into the external validation set for chest pathology detection. 

\textbf{MS-CXR.} The dataset offers phrase grounding annotations that are locally aligned by board-certified radiologists, aiming to support research in the domain of complex semantic modeling for biomedical vision-language tasks \citep{Boecking2022-jw}. Each phrase is associated with at least one bounding box annotated on one CXR. For the region captioning task, we used MS-CXR as the external validation cohort, where box coordinates were passed to the model to generate descriptive text.

\begin{figure}[t]
    \centering
    \includegraphics[width=\textwidth]{Figure7}
    \caption{\textbf{Overview of MedVersa.} \textbf{a,} Processing workflow of MedVersa. Dashed arrows indicate that the associated procedures are contingent upon the decision regarding the utilization of the vision module. Red arrows represent the operations undertaken when employing the vision module. There are three kinds of <Task> in MedVersa: <DET>, <2DSEG>, and <3DSEG>. \textbf{b,} Architecture of the vision-language adapter. \textbf{c,} Illustration of the workflow for chest pathology detection. \textbf{d,} Illustration of the workflow for abdominal organ segmentation. \textbf{e,} Illustration of the workflow for longitudinal study captioning.}
    \label{fig:7}
\end{figure}

\subsection{Pipeline}
As shown in Fig. \ref{fig:7}{\color{Magenta}a}, MedVersa is composed of three components: the multimodal input coordinator, the large language model based optimizable orchestrator, and a variety of learnable vision modules. In practical usage, MedVersa expects inputs in the form of image-request pairs. Note that the vision input may consist of more than one image, which can be multimodal, multiview or multiperiod (see Fig. \ref{fig:banner}). MedVersa autonomously decides whether to use a 2D or a 3D vision encoder to process the vision inputs based on an analysis of the input modality. After receiving the processed inputs, the large language model can decide whether to independently perform the task or utilize a set of visual modeling modules for assistance. This dynamic decision-making process ensures that tasks are handled with the appropriate level of expertise and efficiency. 

\subsection{MedVersa, an orchestrated GMAI system}
\textbf{Multimodal input coordinator.} As Fig. \ref{fig:7}{\color{Magenta}a} displays, the multimodal input coordinator comprises the general vision encoders, the vision-language adapters, and the tokenizer. We design this architecture by taking inspirations from MiniGPT-4 \citep{Zhu2023-tq,Chen2023-nn}, LLaVA-Med \citep{Li2023-nd}, and Med-PaLM M \citep{Tu2023-ef} but keep the architecture easy for implementation. The general vision encoders are the primary gate for vision inputs. Specifically, we exploit distinct encoders for 2D and 3D imaging data, respectively. The 2D vision encoder utilizes the transformer architecture \citep{Vaswani2017-ai} to extract visual tokens from the images. For the 3D encoder, we refer to the encoder from the 3D UNet \citep{Cicek2016-oe}. The extracted visual tokens are concatenated and passed to the adapter to get mapped to the language space. Here, we present an efficient design of the vision-language adapter, which only contains a stack of three layers (see Fig. \ref{fig:7}{\color{Magenta}b}). The first layer is responsible for reducing the number of visual tokens to control the GPU memory cost, which can be achieved with an adaptive pooling function \citep{He2015-wr}. Next, the layer normalization \citep{Ba2016-ef} is applied to the pooled visual tokens, followed by a linear projection layer to map the visual representations to the language space. To align with the 2D and 3D vision encoders, we also employ two independent adapters to process the extracted visual tokens accordingly. Meanwhile, the paired request is processed with the Llama tokenizer \citep{Touvron2023-hp}, which is a byte-pair encoding model based on sentencepiece \citep{Kudo2018-fr}. The request is transformed into a series of textual tokens, which are then contextualized by the following large language model along with the mapped visual tokens. This enables the system to understand and correlate the visual data with the relevant requests.

\textbf{Orchestrated modeling.} Unlike prior research that depended exclusively on large language models (LLMs) for task execution, MedVersa leverages the planning capabilities of the large language model to act as an optimizable orchestrator of system operations. Specifically, the orchestrator has to decide whether to carry out the task independently or use a specific vision module for support based on the analysis of visual and linguistic data. This decision-making process can be formulated as: $llm_{\theta}(I, T)\rightarrow(llm_o, s_{o^k})$. I and T denote  the extracted visual and textual tokens, respectively. $llm$ stands for the large language model. $llm_o$ and $s_{o^k}$ represent the outputs of the language model and the $k$th vision module, respectively. For vision-language tasks, MedVersa only adopts the $llm_o$ as the final language response. For vision-centric tasks, the choice of k is determined based on the $llm_o$. Specifically, the $llm$ determines the task type and generates the relevant <Task> in the $llm_o$. There are three kinds of <Task> included in MedVersa: <DET>, <2DSEG>, and <3DSEG>. The predicted <Task> guides the system in selecting the $k$th visual modeling module from the pool, tailored for executing the task described by <Task> (see Fig. \ref{fig:7}{\color{Magenta}a} for more details). Meanwhile, we index the corresponding latent embeddings of <Task> from the output logits of the $llm$. These embeddings gather the information from the input data and help prompt the visual modeling module to complete the desired task. To accomplish this, the indexed latent embeddings can be either passed directly to the visual detection module or integrated with the intermediate features of the visual segmentation modules. We illustrate the orchestration process on three tasks in Fig. \ref{fig:7}, including the chest pathology detection (Fig. \ref{fig:7}{\color{Magenta}c}), the abdominal organ segmentation (Fig. \ref{fig:7}{\color{Magenta}d}), and the longitudinal study captioning (Fig. \ref{fig:7}{\color{Magenta}e}).

\begin{figure}[htp]
    \centering
    \includegraphics[width=\textwidth]{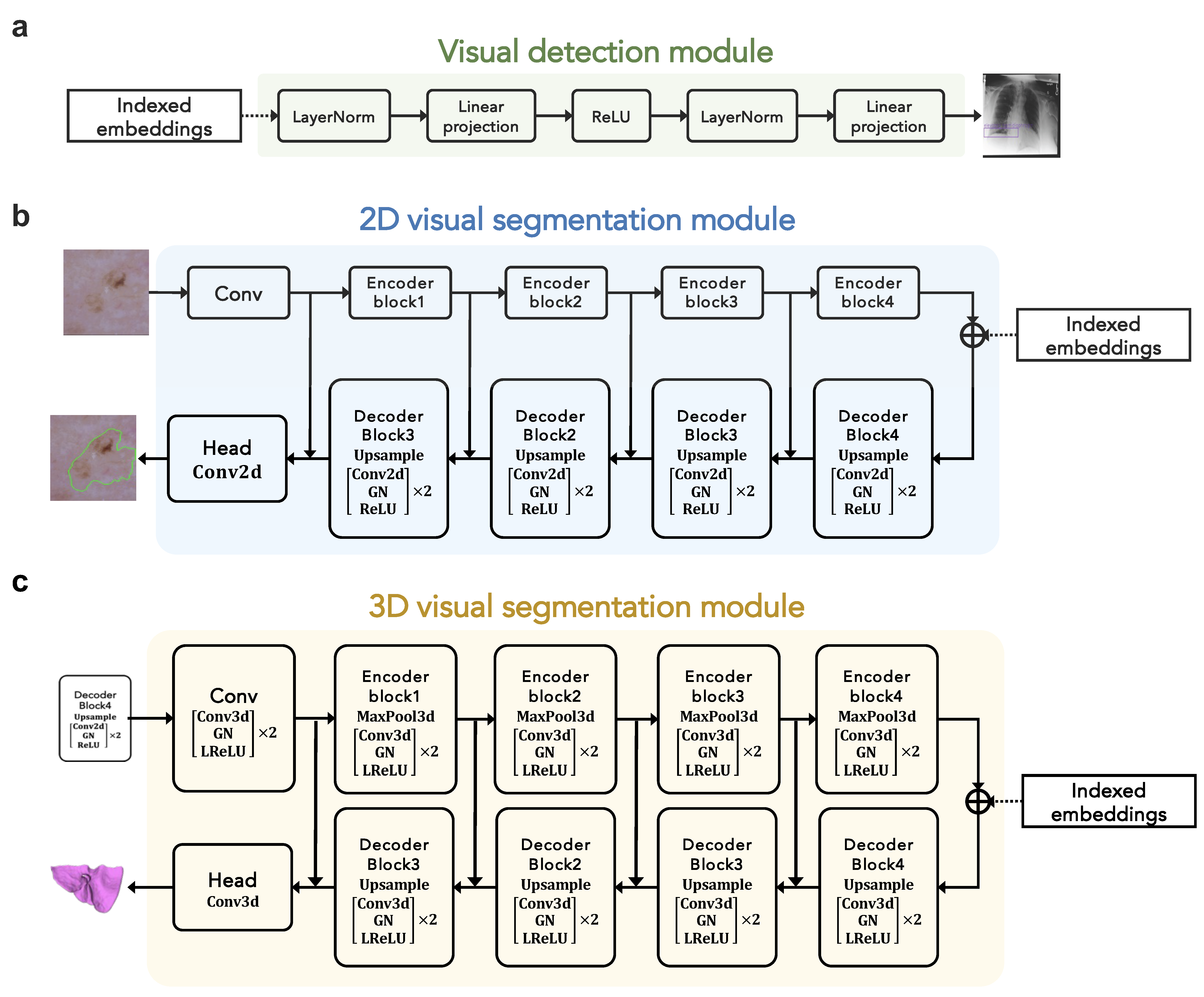}
    \caption{\textbf{Vision modules.} \textbf{a,} Visual detection module. The architecture consists of a sequential neural network that starts with a LayerNorm applied to the hidden size of 4096. This is followed by a linear layer that reduces the dimensionality from 4096 to 256, then a ReLU activation for non-linearity. Another LayerNorm is applied to the 256-dimensional output, followed by a final linear layer that reduces the dimensionality further to 4. \textbf{b,} 2D visual segmentation module. The encoder in this module is built on the ResNet-18 architecture with ImageNet-pretrained weights, while the decoder follows the structure of a UNet decoder. Specifically, we replaced batch normalization layers with group normalization, where we set the number of groups to 16. \textbf{c,} 3D visual segmentation module. In this module, we adopted the UNet3D architecture. Conv2d and Conv3d stand for the 2D and 3D convolution, respectively. GN denotes the group normalization layer, and LReLU represents the leaky ReLU activation function. We initialized the encoder of the specialist module for 2D segmentation using the pretrained weights of ResNet-18 on ImageNet. The red arrows denote the skip connections.}
    \label{fig:8}
\end{figure}

\textbf{Vision modules.} In MedVersa, we have incorporated three modules designed for vision-focused tasks, and these can be readily expanded or replaced if additional or new dedicated modules become necessary. As shown in Fig. \ref{fig:8}{\color{Magenta}a}, we develop a lightweight visual detection module that can be integrated with the orchestrator. For the visual segmentation modules, we employ 2D \citep{Isensee2021-fu} and 3D UNets \citep{Cicek2016-oe} for 2D and 3D image segmentation tasks, respectively (Fig. \ref{fig:8}{\color{Magenta}b} and \ref{fig:8}{\color{Magenta}c}). We initialize the encoder of the 2D UNet using the pretrained weights of ResNet-18 \citep{He2015-xa} on ImageNet \citep{Deng2009-bg}. We have attempted several different approaches to incorporate the indexed embeddings from the large language model into the vision modules. Our observation is that the feature concatenation or addition outperforms the more complex operation, such as cross attention \citep{Jaegle2021-eh}. Based on this, we add the indexed embeddings to the intermediate feature maps in segmentation modules, while feeding these embeddings to the detection module directly. Note that all dedicated modules are optimizable and need to be trained with the other parts of MedVersa.

\textbf{Model training and testing with meticulous, referring image instructions.} The success of Alpaca, along with recent advancements in large language models \citep{Taori2023-ab,Wei2021-vl,Chung2022-en,Singhal2023-ct}, have underscored the importance of incorporating diverse instructions to consolidate multiple tasks and enhance generalization capabilities during supervised fine-tuning. This compelling evidence prompted us to embrace this concept within MedVersa. We propose referring image instruction tuning, where image identifiers are added to instructions to specify different images. This technique enhances the model's capability to perform complex comparative analyses, such as the longitudinal study captioning, where we need to assign images to different studies and compare studies instead of images. 

For example, in Fig. \ref{fig:7}{\color{Magenta}e}, the exact input to MedVersa for longitudinal study captioning is like: ‘<img0>$v_0$</img0><img1>$v_1$</img1><img2>$v_2$</img2><img3>$v_3$</img3> Highlight any difference in <img0><img1> compared to the prior study <img2><img3>.’ $v_i$ stands for the visual tokens of the $i$th input image. We showcase all instructions used in the model training in Table \ref{tab:5}. For each task in our study, we asked ChatGPT to generate a maximum of 20 prompts, each adhering to a predefined template. This template consists of the initial instruction for each task, providing a structured starting point for the prompts. After this, we conducted a manual review, carefully sifting through the generated prompts to eliminate any that were similar in nature, ensuring that we retained only the most diverse and distinct prompts for our analysis. During the training and test phases, for a given sample corresponding to a specific task, we choose an instruction at random from the set of instructions linked to the task. Here, we outline the method for creating ground truth labels for various tasks and samples. For vision-language tasks, natural language answers are directly utilized as the target for training the model. In particular, for classification tasks, the model is instructed to produce the names of diagnoses. When dealing with vision-centric tasks, we employ distinct labeling techniques for detection and segmentation. In detection tasks, during each training iteration, we initially select up to nine classes at random and convey their names (together with a randomly chosen instruction) to the model. The model is trained to append either <N/A> or <DET> tags following each class name. <N/A> indicates the absence of the corresponding class in the input image, whereas <DET> signifies its presence. Subsequently, we identify the latent embeddings of <DET> tags and use them in the visual detection module for determining bounding box coordinates. For segmentation tasks, a single class name is randomly chosen from the set, and this name, along with the instruction, is fed into the model. The model is then trained to generate either ‘The segmentation mask of [class name] is <2DSEG>’ or ‘The segmentation mask of [class name] is <3DSEG>,’ depending on whether the task is 2D or 3D segmentation. A similar idea was also adopted in natural image processing \citep{Lai2023-gr}. As in detection, the relevant embeddings for <2DSEG> or <3DSEG> are then passed to the appropriate 2D or 3D visual segmentation modules to create the segmentation masks.

\textbf{Domain-aware minibatch gradient descent for multimodal multitask training.} Unlike current medical FMs \citep{Tu2023-ef, Wu2023-aa, Moor2023-iy, Huang2023-kf} that only probed the vision-language capability, MedVersa needs to be trained on both vision-language and vision-centric tasks, which brings challenges to classic minibatch gradient descent optimization \citep{Hinton2012-yj}. To solve this, we introduce a method called domain-aware minibatch gradient descent. The main idea is to create minibatches using training samples that all come from the same task and imaging type. We first divide the training data into seven groups based on their tasks: image captioning, classification, detection, segmentation, VQA, region captioning, and longitudinal study captioning. After that, we create minibatches for each group by randomly selecting samples that match the same imaging type. This means that each minibatch will have data focused on one specific task and one type of imaging. For example, one minibatch might include only samples for the segmentation task on CT scans, while another might contain only samples for the detection task on CXRs. During each training iteration, we start by randomly picking an imaging type. Then, we select a task associated with this imaging type and randomly sample data for that task. We apply gradient descent separately to each minibatch, allowing the model to better learn and improve its performance for each specific task and imaging type. We also use different loss functions tailored to each task. For example, cross-entropy loss is used for vision-language tasks, while a mix of cross-entropy and regression losses is used for detection. For the segmentation task, we use both focal loss and DICE loss, giving them equal importance.

\begin{figure}[t]
    \centering
    \includegraphics[width=\textwidth]{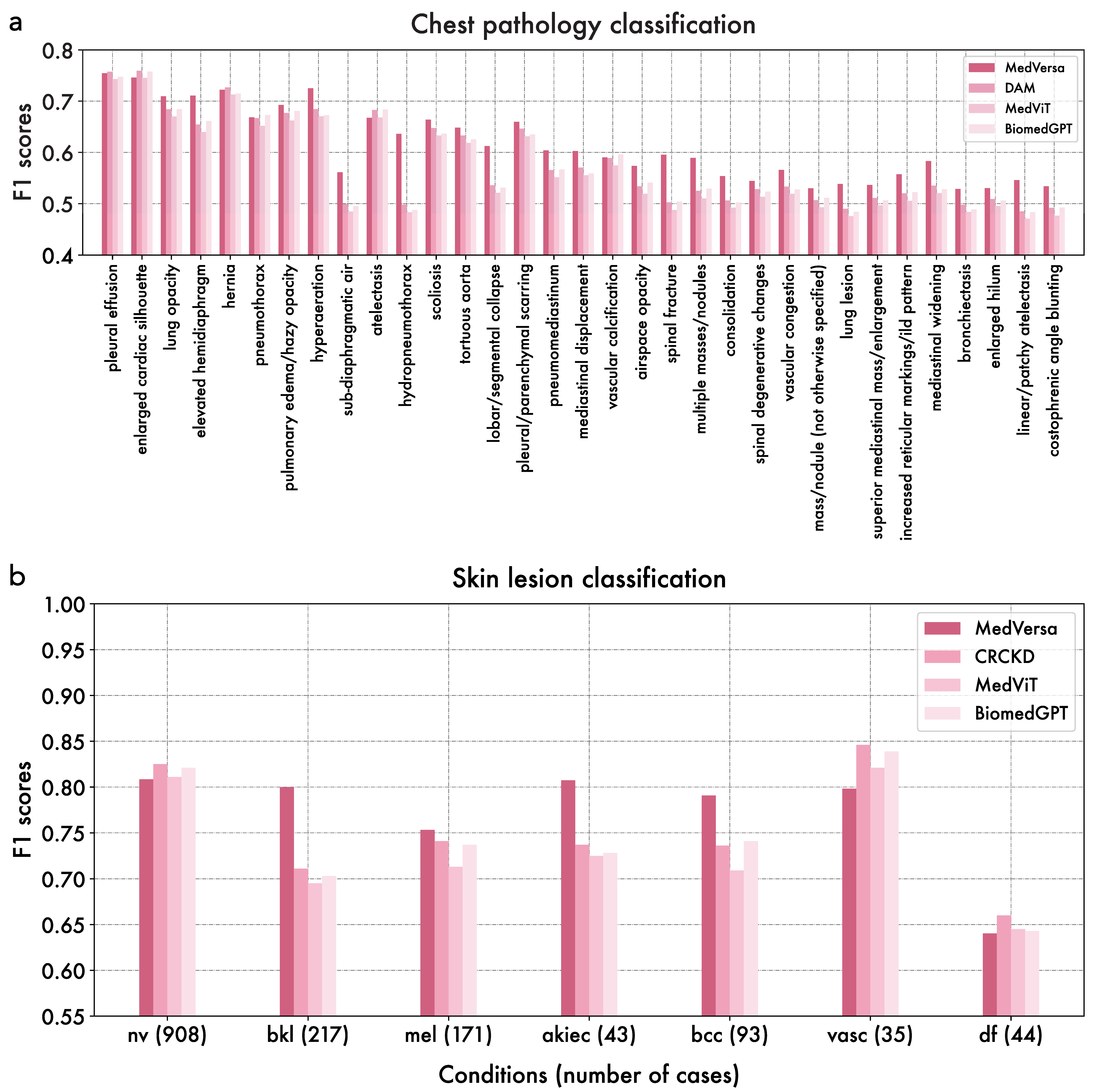}
    \caption{\textbf{Per-category results for image classification.} \textbf{a,} chest pathology classification. \textbf{b,} skin lesion classification.}
    \label{fig:10}
\end{figure}

\textbf{Radiologist evaluation.} A structured radiologist evaluation was conducted using a private instance of Argilla (version 1.29). Radiologists reviewed pairs of reports (one AI-generated by MedVersa, one human-generated) alongside corresponding images and clinical information. Reports were randomly shuffled and labeled as “Report A” and “Report B” for blind evaluation. Radiologists judged whether Report A was better, Report B was better, or both were clinically equivalent. Studies for evaluation were randomly sampled from the test set of MIMIC-CXR.

\textbf{User study.} We conducted a reader study to evaluate artificial intelligence assistance in chest radiograph reporting, recruiting ten board-certified radiologists through a specialized medical image annotation company in India. The study utilized randomly selected chest radiographs from the MIMIC-CXR dataset. We developed a custom evaluation platform that integrated several key components: a DICOM viewer for image interpretation, a worklist interface for case management, and a report editing interface where radiologists could modify and finalize their reports. Upon report completion, the platform automatically prompted radiologists to rate their confidence level and the mental demand required for the case using Likert scales. The platform's integrated timer tracked time spent on each case. The preparation phase included providing participants with a detailed study task protocol and an instructional video, completing a standardized training phase of 25 cases to ensure platform familiarity, and conducting an interactive orientation meeting where we demonstrated platform functionality, reviewed task requirements, and addressed any remaining questions about the study procedures.

For the main evaluation, each radiologist interpreted 75 unique chest radiographs under three different reporting scenarios: (1) starting with a standard negative template, (2) starting with a GPT-4o-generated report draft, or (3) starting with a MedVersa-generated report draft. Cases were randomly and evenly distributed across these three scenarios, with each radiologist reading 25 cases per scenario. For each case, radiologists were tasked with reviewing the chest radiograph and modifying the provided template or draft to produce a clinically accurate final report that would meet the standards of clinical practice. To eliminate potential recall bias and need for washout periods, each radiologist interpreted a case only once, while the same case was interpreted under different scenarios across different radiologists, ensuring balanced distribution of case complexity.

We evaluated three key dimensions of reporting performance: efficiency, radiologist experience, and clinical consistency. Reporting efficiency was measured through the platform's integrated timer. Radiologist experience was assessed through self-reported confidence levels and mental demand ratings collected immediately after each report completion. For clinical consistency, we compared the “findings” section of each final report against those of the other nine radiologists for the same case, using an adapted version of FineRadScore (using GPT-4o) \citep{Huang2024-vi}. We calculated the average number of urgent or emergent discrepancies based on FineRadScore clinical severity categories.

\begin{figure}[t]
    \centering
    \includegraphics[width=\textwidth]{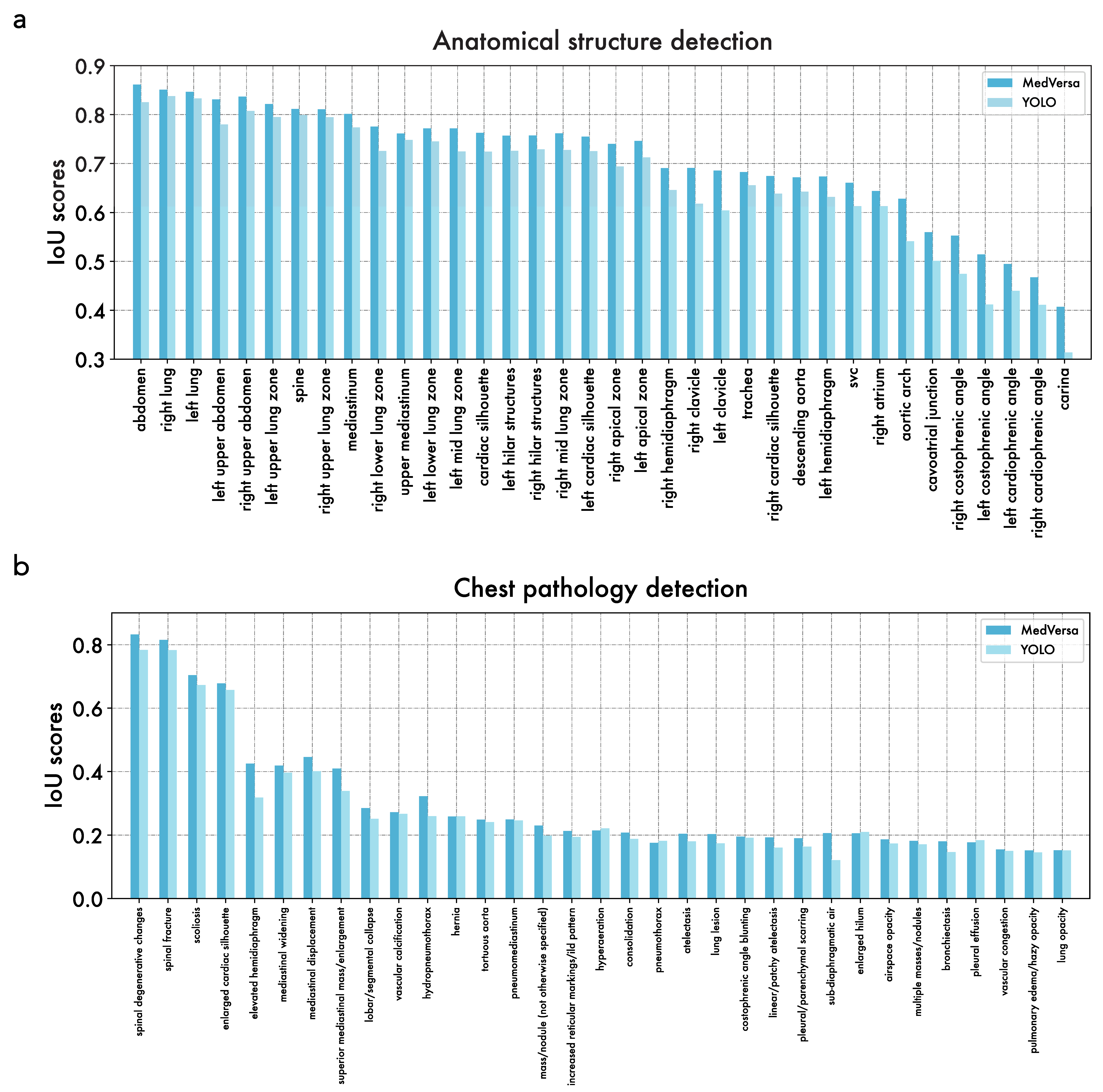}
    \caption{\textbf{Per-category results for object detection.} \textbf{a,} Anatomical structure detection. \textbf{b,} Chest pathology detection.}
    \label{fig:11}
\end{figure}

\subsection{Implementation details}
For the 2D vision encoder in the multimodal input coordinator, we use the base version of Swin Transformer \citep{Liu2021-dq} pretrained on ImageNet \citep{Deng2009-bg}. This encoder is characterized by its four-stage structure, a window size of seven, a patch size of four, and an initial feature dimension of 128. For the 3D vision encoder, we adopt the encoder architecture from the 3D UNet \citep{Cicek2016-oe}. For specific tasks like report generation, classification, open-ended VQA, and longitudinal study captioning, the encoder processes the input images through a random cropping technique, where the cropped area ranges from 50\% to 100\% of the original image. These cropped images are then resized to a standard dimension of 224$\times$224 pixels with three channels. Different augmentation techniques are applied based on the nature of the task. For chest organ and skin lesion segmentation tasks, a random horizontal flip is applied to each image. In the case of abdomen CT scans, a more complex manipulation is performed by flipping each 3D volume over a random axis. To efficiently manage the volume of visual tokens, MedVersa utilizes an adaptive average pooling strategy, standardizing the output length to nine. Additionally, the system implements two distinct linear projectors for 2D and 3D data. Each projector comprises a fully connected layer, transforming each pooled visual token into a 1D vector of 4,096 elements. 

We initialized the LLM-based orchestrator using the model weights of Llama-2-Chat \citep{Touvron2023-hp}. The training of the orchestrator in MedVersa employs the Low-Rank Adaptation (LoRA) strategy \citep{Hu2021-fu} as we found that it outperformed full-parameter training. LoRA utilizes the concept of low-rank matrix decomposition to approximate a large weight matrix in neural network layers. By setting the rank and alpha values of LoRA to 16, the method ensures efficient training while modifying only a fraction of the model parameters. The AdamW optimizer \citep{Loshchilov2017-dx}, in combination with a cosine learning rate scheduler, is used for optimization. Training parameters are meticulously set, with an initial learning rate of 3e-4 and a minimum of 3e-6, over 500,000 training iterations. The first 3,000 iterations involve a linear warm-up phase, starting with a learning rate of 1e-7. Finally, the training infrastructure comprises 24 NVIDIA A100 GPUs (80G). This setup allows the training stage to be completed within a 72-hour window. Following this round of training, we extended MedVersa's segmentation functionality by further finetuning the model on additional images and masks. Specifically, we froze the language model and focused on finetuning the segmentation vision module. In this phase, we applied random horizontal flip as the primary data augmentation strategy. We used an initial learning rate of 1e-4 and finetuned the model for 300 epochs, with an input image size of 224$\times$224.

\begin{figure}[t]
    \centering
    \includegraphics[width=\textwidth]{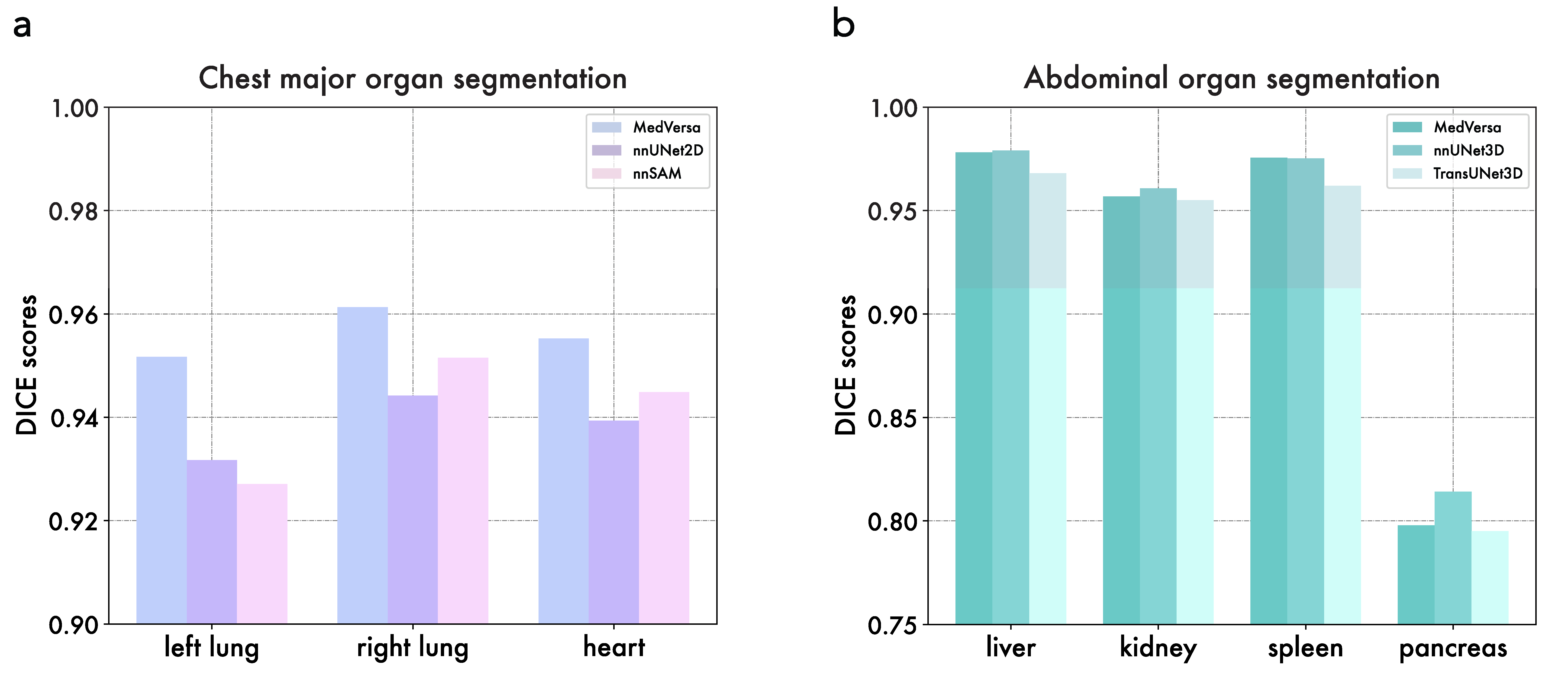}
    \caption{\textbf{Per-category results for image segmentation.} \textbf{a,} Chest major organ segmentation (2D). \textbf{b,} Abdominal organ segmentation (3D).}
    \label{fig:12}
\end{figure}
We use F1 score instead of AUC (Area Under the Curve) for evaluating classification tasks because F1 score better captures the model's performance on both recall and precision. In contrast, AUC summarizes performance across all possible thresholds for classifying an instance as positive. It doesn't emphasize the model's performance at the high-precision, high-recall region. In disease classification tasks, it is crucial to have a high F1 score because false negatives (failing to identify a patient with the disease) and false positives (incorrectly diagnosing a healthy patient with the disease) can both have serious consequences. Therefore, the F1 score is used in our experiments as it balances both precision and recall, ensuring that the model correctly identifies a high proportion of true disease cases while minimizing misdiagnoses. In practice, we found a default threshold 0.5 is sufficient. Considering the robustness and generalization to new, unseen data, we use this threshold when computing precision and recall scores.

\subsection{Baselines}
\begin{enumerate}
    \item \textbf{ClsGen.} This is a differentiable end-to-end method with three parts: a classifier, a generator, and an interpreter \citep{Nguyen2021-gq}. The classifier learns disease features through context modeling and a disease-state aware mechanism. The generator turns the disease information into a medical report. The interpreter then reviews and refines these reports, ensuring they align with the classifier's findings. We empirically found ClsGen showed more consistent performance compared to popular report generation approaches, such as R2Gen \citep{Chen2020-cd} and M2Trans \citep{Miura2020-sd}. We trained the model for 50 epochs, with a batch size of 64. The initial learning rate is set to 3e-4 and adjusted to 3e-5 at the $25$th epoch. The weight decay is set to 1e-2. The input image size is 336$\times$336, and we only applied random cropping as we did in MedVersa as the main augmentation strategy for report generation. For all other settings we adhered to the original paper \citep{Nguyen2021-gq}.
    \item \textbf{DAM.} This method is particularly relevant for addressing complex classification problems, especially when dealing with imbalanced datasets \citep{Yuan2021-zj}. We included the DAM supervised method as a baseline for chest pathology classification, which currently is state-of-the-art on the CheXpert dataset \citep{Tiu2022-zm}. Following the training protocol on CheXpert, we initially trained the DenseNet-121 model \citep{Huang2017-mq}, pretrained on ImageNet, for one epoch using a batch size of 32, a learning rate of 1e-4, and a weight decay of 1e-5. Subsequently, we finetuned the model for two additional epochs, maintaining the same batch size and weight decay, but with an increased learning rate of 0.01. The input image size is 224$\times$224. For other settings, we followed the official tutorial.
    \item \textbf{MAIRA.} This is a specialist large multimodal model for report generation from Microsoft \citep{Hyland2023-al}. It adopted the LLaVA-1.5 architecture \citep{Liu2023-sk}. MAIRA also benefits from the use of GPT-3.5 for data augmentation, adding 131,558 reports with paraphrased findings and indication sections to the training set. MAIRA produces reports with state-of-the-art quality.
    \item \textbf{Med-PaLM M}. This is a large generalist biomedical AI system from Google \citep{Tu2023-ef}. Med-PaLM M was built by finetuning with biomedical data on top of PaLM-E \citep{Driess2023-qw}, a generalist multimodal FM trained on non-medical images and text. Here, we compared to its best variant that has 84 billion parameters, which maintains the state-of-the-art in the task of report generation.
    \item \textbf{PTLM.} It is the state-of-the-art approach on open-ended medical visual question answering \citep{Van_Sonsbeek2023-ni}. PTLM maps the extracted visual features to a set of learnable tokens, which can directly prompt the language model for parameter-efficient finetuning. We utilized 600 warmup steps with a learning rate of 5e-3 and a cosine learning rate scheduler. The model was trained for 30 epochs with a batch size of 64. The input image size is 224$\times$224. For all other configurations, we followed the settings from the official repository.
    \item \textbf{EKAID.} EKAID integrates the expert knowledge graphs into representation learning \citep{Hu2023-yf}. This is an image-difference model that is sensitive to anatomical structures, allowing it to extract image-difference features that are pertinent to the progression of diseases and interventions. EKAID presents state-of-the-art results in the task of longitudinal study captioning. For implementation, we used a learning rate of 0.01 to train the model for 20 epochs at a batch size of 128. The input image size is 224$\times$224, and random cropping is applied to each input image.
    \item \textbf{MiniGPT.} This is a new multimodal foundation model that can caption bounding boxes on natural images \citep{Chen2023-nn}. Specifically, it accepts box coordinates as inputs and outputs a caption that describes the objects within the box. We therefore finetuned MiniGPT on the region captioning task. We used a cosine learning rate scheduler to train the model with a batch size of 16 and a maximum learning rate of 3e-4. The input image size is 224$\times$224, with random cropping applied to each image.
    \item \textbf{CRCKD.} This approach aims to bring similar image pairs from the same skin class closer together in both teacher and student models while pushing apart dissimilar image pairs from different skin classes \citep{Xing2021-xp}. It is a widely adopted baseline and shows competitive performance for categorizing skin lesions. Note that both CRCKD and MedVersa were trained directly on the raw, imbalanced HAM10000 dataset. The input image size is 224$\times$224. We trained the model for 80 epochs with the batch size and the warm-up epoch set as 64 and 30, respectively. The initial learning rate was set to 1e-4 and decayed by the one-cycle schedule. The temperature is set as 0.07. For other settings, we followed the official code repository.
    \item \textbf{YOLO.} YOLO is a state-of-the-art, real-time object detection algorithm that is part of the YOLO (You Only Look Once) family \citep{Jocher2020-xm}. Following its predecessors in providing fast and accurate object detection capabilities, YOLO has been widely used for detecting abnormalities in medical images \citep{Mohiyuddin2022-la,Wan2021-qm,Luo2021-dm}. We trained the model for 100 epochs with an input image size of 224$\times$224. The batch size is 64, and the initial learning rate is 0.01. We set the weight decay to 5e-4 and used the default mosaic augmentation.
    \item \textbf{nnUNet.} nnUNet is a self-configuring method for deep learning-based biomedical image segmentation \citep{Isensee2021-fu}. This framework is versatile in handling various medical imaging datasets, employing different configurations and preprocessing steps depending on the dataset characteristics. It adapts the network topologies, such as 2D UNet and 3D UNet, according to the specific requirements of medical segmentation tasks. We adopted nnUNET's automatically configured networks on the datasets as baseline models for 2D and 3D segmentation. During training, we trained the 2D model for 300 epochs and the 3D model for 500 epochs. The batch size is set as 16. The input size of 2D images is 224$\times$224, while for 3D images it is adjusted to 192$\times$192$\times$64. The initial learning rate is 1e-3 with a polynomial decay scheduler. Extensive data augmentation was applied, including rotation, scaling, elastic deformations, and intensity augmentations.
    \item \textbf{nnSAM.} The nnSAM architecture integrates the robust and effective feature extraction abilities of Segment Anything Model \citep{Kirillov2023-ia} with the adaptive configuration strengths of nnUNet \citep{Li2023-at}. This combination maximizes the potential of each model, with SAM providing high-quality feature extraction and nnUNet enabling the system to automatically adjust to the unique demands of each dataset. nnSAM shows state-of-the-art results in the data-efficient segmentation task. In line with the official code repository, we employed the same training parameters and strategies used by nnUNet.
    \item \textbf{BiomedGPT.} BiomedGPT \citep{Zhang2024-ec} is a generalist vision-language foundation model for biomedical tasks. We utilized the pretrained weights of BiomedGPT and fine-tuned the model on each task individually, following the protocol outlined in the original paper. For fine-tuning, we employed the AdamW optimizer with an initial learning rate of 3e-4, along with a cosine learning rate scheduler. The weight decay was set to 1e-5, and the training process lasted for 30 epochs. The input image size is 256$\times$256.
    \item \textbf{MedViT.} MedViT \citep{Manzari2023-sj} is a CNN-Transformer hybrid model designed for robust and efficient medical image classification. In practice, we used the large version of MedViT. We trained MedViT-L for 100 epochs with a batch size of 128. The input images were resized to 224$\times$224 pixels. We used the AdamW optimizer with an initial learning rate of 0.001, which was decayed by a factor of 0.1 at the $50$th and $75$th epochs.
    \item \textbf{3D TransUNet.} 3D TransUNet \citep{Chen2023-al} is an extension of the 2D TransUNet architecture to 3D medical image segmentation. We fixed the input size to 192$\times$192$\times$64, as what we have done to MedVersa, nnUNet, and nnSAM. We used the AdamW optimizer with an initial learning rate of 1e-4 with a cosine learning rate scheduler. The augmentation strategies include random rotation, scaling, flipping, white Gaussian noise, Gaussian blurring, adjusting brightness and contrast, simulation of low resolution, and Gamma transformation.
\end{enumerate}

\subsection{Confidence intervals}
For the estimation of 95\% confidence intervals, non-parametric bootstrap sampling is utilized. This process includes creating 1,000 bootstrap samples from the unseen validation set through random sampling with replacement, with each sample having the same size as the validation set. We then compute the evaluation metric scores for each of these samples. Upon gathering 1,000 scores for the metrics, we organize these scores sequentially. The performance metrics at the 2.5th and 97.5th percentiles are identified and presented as the performance indicators.

\subsection*{Data availability}
All training and validation data is publicly available, links of which are provided in Table \ref{tab:4}.

\subsection*{Code availability}
Code and models can be accessed via \url{https://tinyurl.com/fhnv3be4} (passwd: HYZ0214). 

\subsection*{Author contributions}
P.R. and H.-Y.Z. conceived the study. H.-Y.Z. planned and executed the experiments and
data analysis. S.A. and S.D. interpreted medical report generation results. J.N.A. designed and executed the user study, and analyzed its results. H.-Y.Z. and P.R. drafted the manuscript. All authors provided critical feedback and substantially contributed to the revision of the manuscript. All authors read and approved the manuscript.

\subsection*{Competing interests}
P.R. is co-founder of a2z Radiology AI. J.N.A is a part-time employee of a2z Radiology AI. Other authors declare no competing interests.

\begin{table}[!htp]
\centering
\caption{\textbf{Overview of MedInterp.} All datasets included in MedInterp are publicly available. We reported the dataset size after preprocessing. For each dataset, we also denoted the associated task(s) and the stage(s) involved. VQA denotes visual question answering. 1, 2, 3 in the stages column stand for the training, internal validation, and external validation stages, respectively. We followed the guideline in SAM-Med to preprocess the segmentation data \citep{Cheng2023-gq}.}
\resizebox{\textwidth}{!}{
\begin{tabular}{l|l|l|l|l}
\toprule
\textbf{Datasets} & \textbf{Size} & \textbf{Imaging modalities} & \textbf{Tasks} & \textbf{Stages} \\ \hline
MIMIC-CXR \citep{Johnson2019-er,Johnson2019-ug} & 216,420 studies & Chest X-ray & Image captioning & 1, 2 \\ \hline
\multirow{4}{*}{Chest ImaGenome \citep{Wu2021-fp}} & 235,721 images & Chest X-ray & Image classification & 1, 2 \\ \cline{2-5}
 & 8,425,163 boxes & Chest X-ray & \makecell[l]{Object detection \\ (anatomical structure)} & 1, 2 \\ \cline{2-5}
 & 2,922,665 boxes & Chest X-ray & \makecell[l]{Object detection \\ (pathology)} & 1, 2 \\ \cline{2-5}
 & 2,104,211 captions & Chest X-ray & Region captioning & 1, 2 \\ \hline
\multirow{3}{*}{Medical-Diff-VQA \citep{Hu2023-yf}} & 383,683 QA pairs & Chest X-ray & Open-ended VQA & 1, 2 \\ \cline{2-5}
 & 147,269 comparisons & Chest X-ray & Longitudinal captioning & 1, 2 \\ \cline{2-5}
 & 2,883 QA pairs & Chest X-ray & Open-ended VQA & 3 \\ \hline
PMC-VQA \citep{Zhang2023-el} & 176,946 QA pairs & Mixed & VQA & 1 \\ \hline
PathVQA \citep{He2020-ox} & 32,795 QA pairs & Histopathology & VQA & 1, 2 \\ \hline
VQA-RAD \citep{Lau2018-ow} & 451 QA pairs & Chest X-ray, CT, MRI & VQA & 3 \\ \hline
\multirow{2}{*}{HAM10000 \citep{Tschandl2018-zi}} & 11,526 images & Dermoscopy & Image classification & 1, 2 \\ \cline{2-5}
 & 10,015 masks & Dermoscopy & 2D image segmentation & 1, 2 \\ \hline
CheXpert \citep{Irvin2019-hd} & 668 images & Chest X-ray & Image classification & 3 \\ \hline
IUX-ray \citep{Demner-Fushman2016-kk} & 3,323 studies & Chest X-ray & Image captioning & 3 \\ \hline
NIH ChestX-ray \citep{Wang2017-sg} & 577 boxes & Chest X-ray & Object detection & 3 \\ \hline
MS-CXR \citep{Boecking2022-jw} & 1,448 captions & Chest X-ray & Region captioning & 3 \\ \hline
AbdomenCT-1K \citep{Ma2021-vq} & 3,964 masks & CT & 3D image segmentation & 1, 2 \\ \hline
\multirow{2}{*}{CheXmask \citep{gaggion2024chexmask}} & 719,793 masks & Chest X-ray & 2D image segmentation & 1, 2 \\ \cline{2-5}
 & 600 masks & Chest X-ray & 2D image segmentation & 2 \\ \hline
ACDC \citep{Bernard2018-en} & 4,976 masks & MR & 2D image segmentation & 1, 2 \\ \hline
AMOS2022 \citep{Ji2022-lm} & 217,383 masks & CT, MR & 2D image segmentation & 1, 2 \\ \hline
ASC18 \citep{Xiong2021-sy} & 8,855 masks & MR & 2D image segmentation & 1, 2 \\ \hline
ATM2022 \citep{Zhang2023-za} & 41,604 masks & CT & 2D image segmentation & 1, 2 \\ \hline
AbdomenCT1K \citep{Ma2022-zf} & 217,155 masks & CT & 2D image segmentation & 1, 2 \\ \hline
BTCV \citep{Landman2015-qn} & 10,243 masks & CT & 2D image segmentation & 1, 2 \\ \hline
BTCV\_Cervix \citep{Landman2015-qn} & 4,667 masks & CT & 2D image segmentation & 1, 2 \\ \hline
BraTS2013 \citep{Menze2015-wr,Kistler2013-xn} & 118,496 masks & MR & 2D image segmentation & 1, 2 \\ \hline
BraTS2015 \citep{Menze2015-wr,Kistler2013-xn} & 615,336 masks & MR & 2D image segmentation & 1, 2 \\ \hline
BraTS2018 \citep{Menze2015-wr,Bakas2017-qv,Bakas2018-eo} & 537,300 masks & MR & 2D image segmentation & 1, 2 \\ \hline
BraTS2019 \citep{Menze2015-wr,Bakas2017-qv,Bakas2018-eo} & 629,196 masks & MR & 2D image segmentation & 1, 2 \\ \hline
BraTS2020 \citep{Menze2015-wr,Bakas2017-qv,Bakas2018-eo} & 699,956 masks & MR & 2D image segmentation & 1, 2 \\ \hline
BraTS2021 \citep{Menze2015-wr,Bakas2017-qv,Bakas2018-eo} & 1,255,832 masks & MR & 2D image segmentation & 1, 2 \\ \hline
BrainTumour \citep{Nickparvar2021-oe} & 991,171 masks & MR & 2D image segmentation & 1, 2 \\ \hline
Brain\_PTM \citep{Nelkenbaum2020-kd,Avital2020-lk} & 12,698 masks & MR & 2D image segmentation & 1, 2 \\ \hline
CAD\_PE \citep{Gonzalez2020-ju} & 8,307 masks & CT & 2D image segmentation & 1, 2 \\ \hline
CHAOS\_Task\_4 \citep{Kavur2021-pi} & 3,513 masks & MR & 2D image segmentation & 1, 2 \\ \hline
CMRxMotions \citep{Wang2022-yb} & 3,312 masks & MR & 2D image segmentation & 1, 2 \\ \hline
COSMOS2022 \citep{Antonelli2022-xr} & 2,530 masks & MR & 2D image segmentation & 1, 2 \\ \hline
COVID-19-20 \citep{Roth2022-sz} & 7,321 masks & CT & 2D image segmentation & 1, 2 \\ \hline
COVID19CTscans \citep{Ma2021-vq} & 18,880 masks & CT & 2D image segmentation & 1, 2 \\ \hline
CTPelvic1k \citep{Liu2020-il} & 551,693 masks & CT & 2D image segmentation & 1, 2 \\ \hline
CTSpine1k\_Full \citep{Deng2021-mg} & 419,191 masks & CT & 2D image segmentation & 1, 2 \\ \hline
CT\_ORG \citep{Rister2020-sn} & 599,292 masks & CT & 2D image segmentation & 1, 2 \\ \hline
CrossMoDA21 \citep{Dorent2022-dh} & 739 masks & MR & 2D image segmentation & 1, 2 \\ \hline
CrossMoDA22 \citep{Shusharina2021-wd} & 1,478 masks & MR & 2D image segmentation & 1, 2 \\ \hline
EMIDEC \citep{Lalande2022-xw} & 1,694 masks & MR & 2D image segmentation & 1, 2 \\ \hline
FLARE21 \citep{Ma2022-to} & 129,239 masks & CT & 2D image segmentation & 1, 2 \\ \hline
FLARE22 \citep{Ma2023-il} & 23,368 masks & CT & 2D image segmentation & 1, 2 \\ \hline
Heart\_Seg\_MRI \citep{Tobon-Gomez2015-ab} & 517 masks & MR & 2D image segmentation & 1, 2 \\ \hline
ISLES2016 \citep{Winzeck2018-oy} & 1,506 masks & MR & 2D image segmentation & 1, 2 \\ \hline
ISLES2017 \citep{Winzeck2018-oy} & 2,022 masks & MR & 2D image segmentation & 1, 2 \\ \hline
ISLES2018 \citep{Hakim2021-gq,Cereda2016-iz} & 1,635 masks & CT & 2D image segmentation & 1, 2 \\ \hline
ISLES2022 \citep{Hernandez-Petzsche2022-ia} & 13,572 masks & MR & 2D image segmentation & 1, 2 \\ \hline
ISLES\_SISS \citep{Maier2017-sy} & 15,648 masks & MR & 2D image segmentation & 1, 2 \\ \hline
ISLES\_SPES \citep{Maier2017-sy} & 24,066 masks & MR & 2D image segmentation & 1, 2 \\ \hline
Instance22 \citep{Li2022-fb,Li2023-gi} & 920 masks & CT & 2D image segmentation & 1, 2 \\ \hline
KiTS2019 \citep{Heller2019-wx} & 54,726 masks & CT & 2D image segmentation & 1, 2 \\ \hline
KiTS2021 \citep{Zhao2022-nq} & 80,540 masks & CT & 2D image segmentation & 1, 2 \\ \hline
LNDb \citep{Pedrosa2019-gj} & 1,010 masks & CT & 2D image segmentation & 1, 2 \\ \hline
LUNA16 \citep{Setio2017-oq} & 367,732 masks & CT & 2D image segmentation & 1, 2 \\ \hline
LongitudinalMultipleSclerosisLesionSegmentation \citep{Carass2017-yc} & 13,372 masks & MR & 2D image segmentation & 1, 2 \\ \hline
MMWHS \citep{Zhang2019-kn,Zhuang2016-rl,Luo2023-pa} & 42,682 masks & CT & 2D image segmentation & 1, 2 \\ \bottomrule
\end{tabular}}
\label{tab:4}
\end{table}

\begin{table}[t]
\ContinuedFloat
\centering
\resizebox{\textwidth}{!}{
\begin{tabular}{l|l|l|l|l}
\toprule
MSD\_Colon \citep{Antonelli2022-xr} & 1,197 masks & CT & 2D image segmentation & 1, 2 \\ \hline
MSD\_Heart \citep{Antonelli2022-xr} & 1,101 masks & MR & 2D image segmentation & 1, 2 \\ \hline
MSD\_HepaticVes \citep{Antonelli2022-xr} & 6,576 masks & CT & 2D image segmentation & 1, 2 \\ \hline
MSD\_Liver \citep{Antonelli2022-xr} & 82,784 masks & CT & 2D image segmentation & 1, 2 \\ \hline
MSD\_Lung \citep{Antonelli2022-xr} & 2,347 masks & CT & 2D image segmentation & 1, 2 \\ \hline
MSD\_Pancreas \citep{Antonelli2022-xr} & 10,349 masks & CT & 2D image segmentation & 1, 2 \\ \hline
MSD\_Prostate \citep{Antonelli2022-xr} & 1,466 masks & MR & 2D image segmentation & 1, 2 \\ \hline
MSD\_Spleen \citep{Antonelli2022-xr} & 1,008 masks & CT & 2D image segmentation & 1, 2 \\ \hline
PALM19 \citep{Fang2024-uk} & 1,144 masks & Fundus & 2D image segmentation & 1, 2 \\ \hline
PROMISE12 \citep{Litjens2014-fk} & 776 masks & MR & 2D image segmentation & 1, 2 \\ \hline
Parse22 \citep{Luo2023-pa} & 48,225 masks & CT & 2D image segmentation & 1, 2 \\ \hline
Promise09 \citep{Bharatha2001-jl} & 148 masks & MR & 2D image segmentation & 1, 2 \\ \hline
Prostate\_MRI\_Segmentation\_Dataset \citep{Liu2020-gt} & 1,865 masks & MR & 2D image segmentation & 1, 2 \\ \hline
StructSeg2019\_subtask1 \citep{Antonelli2022-xr} & 3,673 masks & CT & 2D image segmentation & 1, 2 \\ \hline
StructSeg2019\_subtask2 \citep{Antonelli2022-xr} & 4,772 masks & CT & 2D image segmentation & 1, 2 \\ \hline
Totalsegmentator\_dataset \citep{Wasserthal2023-os} & 5,520,406 masks & CT & 2D image segmentation & 1, 2 \\ \hline
VESSEL2012 \citep{Rudyanto2014-qy} & 33,784 masks & CT & 2D image segmentation & 1, 2 \\ \hline
VerSe19 \citep{Sekuboyina2021-bj} & 208,016 masks & CT & 2D image segmentation & 1, 2 \\ \hline
VerSe20 \citep{Sekuboyina2021-bj} & 314,067 masks & CT & 2D image segmentation & 1, 2 \\ \hline
WORD \citep{Luo2022-au} & 87,346 masks & CT & 2D image segmentation & 1, 2 \\ \hline
autoPET \citep{Gatidis2022-wg} & 14,457 masks & CT & 2D image segmentation & 1, 2 \\ \hline
brainMRI \citep{Antonelli2022-xr} & 1,441 masks & MR & 2D image segmentation & 1, 2 \\ \hline
Breast\_ultrasound\_images\_dataset \citep{Al-Dhabyani2020-fi} & 645 masks & Ultrasound & 2D image segmentation & 1, 2 \\ \hline
Cranium \citep{Hssayeni2020-ee} & 229 masks & CT & 2D image segmentation & 1, 2 \\ \hline
cvc\_clinicdb \citep{Bernal2015-kr} & 645 masks & Endoscopy & 2D image segmentation & 1, 2 \\ \hline
Endovis15 \citep{Bernal2017-ix} & 645 masks & Endoscopy & 2D image segmentation & 1, 2 \\ \hline
Gamma \citep{Fu2018-zn,Fu2020-bg} & 190 masks & Fundus & 2D image segmentation & 1, 2 \\ \hline
Hvsmr\_2016 \citep{Pace2015-zz} & 7,479 masks & MR & 2D image segmentation & 1, 2 \\ \hline
Ichallenge\_adam\_task2 \citep{Fang2022-ak} & 270 masks & Fundus & 2D image segmentation & 1, 2 \\ \hline
Kvasir\_seg \citep{Jha2020-zj} & 211 masks & Endoscopy & 2D image segmentation & 1, 2 \\ \hline
Kvasircapsule\_seg \citep{Jha2021-py} & 55 masks & Endoscopy & 2D image segmentation & 1, 2 \\ \hline
M\&Ms-2 \citep{Campello2021-gl} & 6,782 masks & MR & 2D image segmentation & 1, 2 \\ \hline
PI-CAI \citep{Saha2024-qa} & 451 masks & MR & 2D image segmentation & 1, 2 \\ \hline
PI-CAI\_semi \citep{Saha2024-qa} & 620 masks & MR & 2D image segmentation & 1, 2 \\ \hline
Ultrasound Nerve Segmentation \citep{Antonelli2022-xr} & 2,323 masks & Ultrasound & 2D image segmentation & 1, 2 \\ \bottomrule
\end{tabular}}
\end{table}
\clearpage

\begin{table}[htp]
\centering
\caption{\textbf{Instructions used by MedVersa in different tasks.} \_*\_ is the placeholder for the main image identifier (e.g., <img0>). -*- denotes the placeholder for the abnormalities, bounding box coordinates, reference image identifier, detection, and segmentation targets in binary classification, region captioning, longitudinal captioning, detection, and segmentation tasks, respectively. For each task, we asked ChatGPT to generate at most 20 prompts based on a predefined template (i.e., the first instruction for each task. Then, we manually filtered out similar prompts and kept those diverse ones.}
\resizebox{\textwidth}{!}{
\begin{tabular}{ll}
\toprule
\textbf{Tasks} & \textbf{Instructions} \\ \hline
Image captioning (report generation, findings section) & \makecell[l]{
1. Can you detail the findings observed in \_*\_? \\
2. Kindly enumerate the findings from \_*\_. \\
3. I'd like a breakdown of the findings from \_*\_. \\
4. I'd like a section on the findings derived from \_*\_. \\
5. Please write a finding section for \_*\_. \\
6. Would you please write a finding section for \_*\_? \\
7. Please write a section of findings for \_*\_. \\
8. Would you please write a section of findings for \_*\_? \\
9. How would you characterize the findings from \_*\_? \\
10. Please list the discernible findings from \_*\_. \\
11. Can you compile a list of all the notable findings present in \_*\_? \\
12. Please document any findings you see in \_*\_. } \\ \hline
Image captioning (report generation, impression section) & \makecell[l]{
1. Can you please provide your overall impression of \_*\_? \\
2. What's your main impression from \_*\_? \\
3. Please draft a concise impression on \_*\_. \\
4. Would you give a comprehensive impression based on \_*\_? \\
5. I'm looking for an impression for \_*\_. \\
6. Provide your diagnostic impression based on the \_*\_. \\
7. Draft an impression for \_*\_. \\
8. Would you please write an impression section for \_*\_? \\
9. Summarize the impression for \_*\_. }\\ \hline
Image captioning
(report generation, complete report) & \makecell[l]{1. Can you provide a radiology report for \_*\_?\\
2. Please report \_*\_.\\
3. Can you provide a report of \_*\_ with findings and impression?\\
4. Report \_*\_ with findings and impression.\\
5. Please write a radiology report for \_*\_.\\
6. Please generate a radiology report for \_*\_.\\
7. Please provide a detailed report for \_*\_.\\
8. Can you provide a comprehensive report of \_*\_?\\
9. Please write a radiology report for \_*\_.\\
10. Can you give a thorough report of \_*\_?\\
11. Could you please report \_*\_?\\
12. Can you provide a comprehensive report for \_*\_?\\
}\\ \hline
Image classification
(chest pathology and skin lesion) & \makecell[l]{
1. What is the diagnosis for \_*\_?\\
2. Based on \_*\_, what type of lung disease is suspected?\\
3. Can you identify any abnormality in \_*\_?\\
4. What pathology is indicated by \_*\_?\\
5. What lung disease is likely present in \_*\_?\\
6. What are your conclusions from \_*\_?\\
7. What is your interpretation result of \_*\_?\\
8. What abnormalities are present in \_*\_?\\
9. What is the differential diagnosis for the findings in \_*\_?}\\ \hline
Region captioning & \makecell[l]{
1. Describe region -*- in \_*\_.\\
2. Detail any abnormalities in -*- of \_*\_.\\
3. Can you characterize the features within -*- on \_*\_?\\
4. Please provide an analysis of the anomalies seen in -*- within \_*\_.\\
5. Describe any pathological findings within -*- of \_*\_.\\
6. Highlight and explain any abnormalities you detect in -*- of \_*\_.\\
7. Identify and describe any abnormality in -*- of \_*\_.\\
8. Could you please describe the region -*- in \_*\_?\\
9. Would you please describe the region -*- in \_*\_?\\
10. Give a description of the region -*- in \_*\_.} \\ \hline
Longitudinal study captioning & \makecell[l]{
1. Highlight any difference in \_*\_ compared to the prior study -*-.\\
2. Identify any progression in \_*\_ since the last study -*-.\\
3. Compare the current study \_*\_ with the past one -*-.\\
4. Present any changes in \_*\_ since the last study -*-.\\
5. Detail any progression or regression in \_*\_ in comparison to the older study -*-.\\
6. Detect changes in \_*\_ compared to the past study -*-.\\
7. Compare \_*\_ with the prior study -*- and tell me any difference.} \\ \bottomrule
\end{tabular}}
\label{tab:5}
\end{table}

\begin{table}[t]
\ContinuedFloat
\centering
\resizebox{\textwidth}{!}{
\begin{tabular}{ll}
\toprule
Object detection
(anatomical structure and chest pathology) & \makecell[l]{
1. Detect any signs of -*- in \_*\_.\\
2. Highlight the areas that indicate -*- in \_*\_.\\
3. Show me the regions in \_*\_ where -*- might be present.\\
4. Assess \_*\_ and mark areas consistent with -*- findings.\\
5. Locate and circle any features of -*- in \_*\_.\\
6. Compare \_*\_ to typical -*- patterns and highlight any matches.\\
7. Detect and display potential symptoms of -*- within \_*\_.\\
8. Is there any trace of -*- in \_*\_? Point it out.\\
9. Help me spot -*- by illuminating its markers in \_*\_.\\
10. Search for any characteristic signs of -*- in \_*\_.\\
11. Examine and underscore the presence of -*- in \_*\_.\\
12. Would you please help me locate -*- in \_*\_?\\
13. Could you please help me locate -*- in \_*\_?\\
14. Please help me locate -*- in \_*\_?}\\\hline
Image segmentation & \makecell[l]{
1. Segment -*- in \_*\_.\\
2. Highlight the boundaries of -*- in \_*\_.\\
3. Isolate and show only -*- from \_*\_.\\
4. Can you delineate -*- in \_*\_?\\
5. Segment -*- from the given \_*\_.\\
6. I need a clear segmentation of -*- in \_*\_, please.\\
7. Outline the contours of -*- in \_*\_.\\
8. Show a clear boundary around -*- in \_*\_.\\
9. Separate -*- from the surrounding anatomy in \_*\_.\\
10. Provide a segmented view of -*- in \_*\_.\\
11. Please identify and segment -*- from the rest in \_*\_.\\
12. Give me a clear cutout of -*- in \_*\_.\\
13. Please mask everything except for -*- in \_*\_.\\
14. Draw a boundary around -*- in \_*\_.\\
15. Would you please help me segment -*- in \_*\_?\\
16. Could you please help me segment -*- in \_*\_?\\
17. Please help me segment -*- in \_*\_?} \\
\bottomrule
\end{tabular}}
\end{table}

\clearpage
\newpage
\bibliographystyle{unsrtnat}
\bibliography{refs}
\end{document}